\documentclass[12pt]{iopart}

\usepackage{graphicx}
\usepackage{subcaption}
\usepackage{wrapfig}
\usepackage{booktabs}
\usepackage[usenames,dvipsnames,table]{xcolor}
\usepackage{hyperref}

\usepackage{hyperref}


\begin{document}

\title[]{Machine learning approaches for automatic defect detection in photovoltaic systems}

\author{Swayam Rajat Mohanty}
\address{Department of Electrical Engineering, Indian Institute of Technology Banaras Hindu University, Varanasi 221005, India}
\author{Moin Uddin Maruf}
\address{Department of Mechanical Engineering, Texas Tech University, Lubbock, Texas 79409, USA}
\author{Vaibhav Singh}
\address{Vayv Energy Systems, New Delhi 110044, India}
\author{Zeeshan Ahmad}
\address{Department of Mechanical Engineering, Texas Tech University, Lubbock, Texas 79409, USA}
\ead{zeeahmad@ttu.edu}

\vspace{10pt}

\begin{abstract}
Solar photovoltaic (PV) modules are prone to damage during manufacturing, installation and operation which reduces their power conversion efficiency. This diminishes their positive environmental impact over the lifecycle. Continuous monitoring of PV modules during operation via unmanned aerial vehicles is essential to ensure that defective panels are promptly replaced or repaired to maintain high power conversion efficiencies. Computer vision provides an automatic, non-destructive and cost-effective tool for monitoring defects in large-scale PV plants. We review the current landscape of deep learning-based computer vision techniques used for detecting defects in solar modules. We compare and evaluate the existing approaches at different levels, namely the type of images used, data collection and processing method, deep learning architectures employed, and model interpretability. Most approaches use convolutional neural networks together with data augmentation or generative adversarial network-based techniques. We evaluate the deep learning approaches by performing interpretability analysis on classification tasks. This analysis reveals that the model focuses on the darker regions of the image to perform the classification. We find clear gaps in the existing approaches while also laying out the groundwork for mitigating these challenges when building new models. We conclude with the relevant research gaps that need to be addressed and approaches for progress in this field: integrating geometric deep learning with existing approaches for building more robust and reliable models, leveraging physics-based neural networks that combine domain expertise of physical laws to build more domain-aware deep learning models, and incorporating interpretability as a factor for building models that can be trusted. The review points towards a clear roadmap for making this technology commercially relevant.
\end{abstract}

\section{Introduction}

The widespread adoption of renewable energy technologies is essential for decarbonization while meeting the increasing energy demand. Solar photovoltaic (PV) is expected to dominate the transition to clean energy, accounting for over 65\% of the electricity grid in the future \cite{PVmarket}.  In 2023, the global installed capacity of photovoltaic (PV) systems reached an impressive 1.6 terawatts (TW), with 420 gigawatts (GW) of new installations occurring within the year alone~\cite
{PVmarket}. The solar PV panel market is substantial, valued at approximately 232 billion USD \cite{marketsize}, and the operations and maintenance segment is projected to expand to 15 billion USD by 2030 \cite{saur}.

The lifespan and conversion efficiency of PV systems critically affect the levelized cost of electricity and the environmental benefits of solar energy. To fulfill the promise of a low carbon footprint energy source (41 g CO2 equivalents (CO2e) per kWh of electricity) \cite{Technology}, solar assets must maintain a good performance ratio for at least 25 years. However, underperformance due to difficult-to-spot defective solar cells poses significant financial and environmental challenges.
The defects in solar cells may be introduced during manufacturing or operation by climate conditions such as increased temperature, humidity, hail, winds, and dust. These factors introduce mechanical and thermal strains on the PV modules, leading to rapid degradation and loss of efficiency.
Solar plants operating at low efficiencies or those decommissioned prematurely due to defects can result in an impact of 80 g CO2e per kWh of electricity generated which is substantially higher than the target. This figure contrasts sharply with the more competitive 11 g CO2e per kWh of electricity for wind energy \cite{wind}. %
The true extent of PV module degradation is currently underestimated as most PV plants are less than 10 years old. The issue is expected to worsen with trends showing many plants being decommissioned or renovated ahead of schedule due to inflated maintenance costs or underperformance. Such trends adversely affect the financing and insurance sectors for renewables, already a significant bottleneck for large-scale solar deployment. 

To address the issue of solar panel underperformance, rapid defect and damage assessment of large-scale solar PV plants is essential. Underperformance in solar panels can stem from manufacturing or material-level defects such as microcracks or during operation e.g., shading and soiling. These defects can be detected as distinguishable visual patterns using imaging techniques like thermal infrared (IR), electroluminescence (EL), photoluminescence (PL), and ultraviolet fluorescence (UVF). However, identifying and classifying defects, and locating them in operational solar power plants without downtime or manual intervention, remains a significant challenge. Currently, these processes rely heavily on human effort and are not scalable.

Deep learning-based computer vision techniques, applied to images of solar panels offer the potential of automated defect detection, providing a reliable, cost-effective, and non-invasive solution for solar panel inspection~\cite{bartlerAutomatedDetectionSolar2018,TANG2020453,chenAutomatedDefectIdentification2022,hijjawiReviewAutomatedSolar2023}. These images can be captured by unmanned aerial vehicles (UAVs) equipped with sensors and cameras. These tools, which can be deployed at scale, can provide an effective approach to maintaining the efficiency and longevity of solar PV plants~\cite{Salameh2020LifeCA}.

\begin{figure}[htbp]
    \centering
    \includegraphics[width=\textwidth]{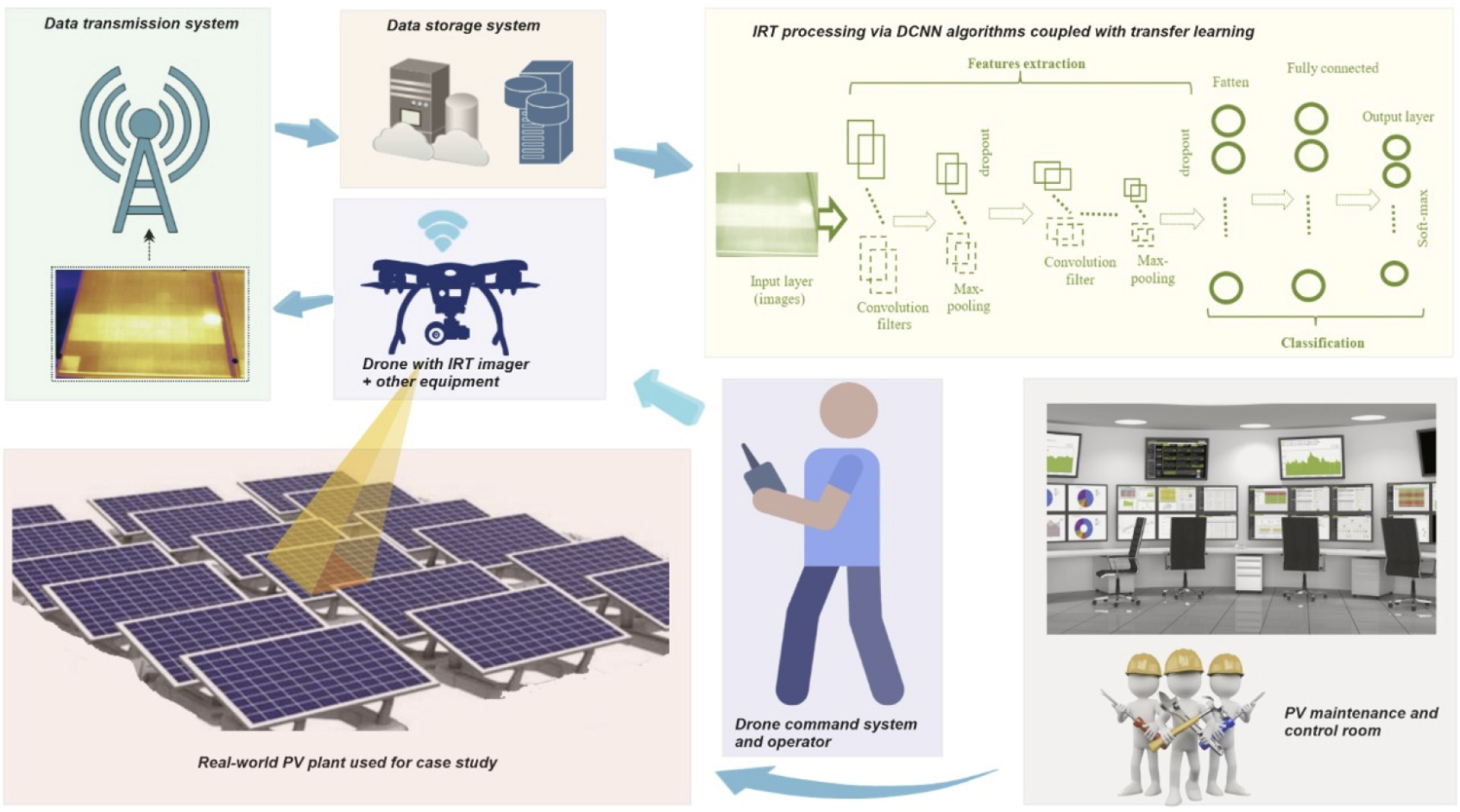}
    \caption{
    An example workflow for automated defect detection in solar panels using unmanned aerial vehicles (UAVs) \cite{rs15061686}. The process begins with the capture of images of solar panels by UAV through manual or automated flight systems. The images are transmitted to a cloud-based data storage platform. This transmission ensures that the images are securely stored and can be accessed remotely. From the cloud storage, the images are retrieved by a computing unit that performs several preprocessing steps on the images to enhance their quality and prepare them for analysis. The computing unit subsequently uses a machine learning model to analyze the images and detect any defects in individual solar cells and modules. This model processes the images and provides inference results, identifying potential issues in the solar panels based on the patterns and anomalies detected in the images.}
    \label{fig:Workflow}
\end{figure}

This review provides a practical overview of the recent advancements in deep learning-based tools and techniques for detecting defects in solar panels. An overview of the workflow involved in defect detection is shown in \autoref{fig:Workflow}. The defect detection process begins with the collection of images through UAVs. The specific imaging technique required depends on the targeted defect. Once collected, the images undergo various preprocessing steps to prepare them to be used as input data for deep learning models. The choice of imaging type, preprocessing methods, and deep learning approaches varies significantly depending on the defect type, requiring a unique combination for accurate detection. The subsequent sections will provide a detailed analysis of how the detection methods differ according to the specific defect type, highlighting the diversity of approaches necessary. We further provide a detailed analysis of the interpretability of the deep learning techniques used to classify defects in solar panels. The interpretability analysis reveals flaws in defect classification and opens up opportunities for further development of techniques to enhance trust in the models. 

\section{Collection of image data}

The initial step of image data collection is a common precursor to most defect detection methods. However, the specific type of image collected significantly impacts the subsequent processing steps and the range of defects that can be identified. Most of the defect detection studies rely on one or more of three types of images: infrared (IR) images, electroluminescence (EL) images, and traditional RGB images.

\subsection{Infrared (IR) images}
These images are obtained by using a thermographic sensor to capture the heat radiated by the solar panel. As the efficiency of the solar cells is temperature-dependent \cite{SINGH201236}, this technique is useful to assess the cell condition. A major cause of losses in PV cells is the degradation of electrical interconnections. Due to this degradation, the defective cells have non-uniform current densities, which causes parts of them to heat up and appear as a hotspots in the IR image. On the other hand, a shunt in the cell appears as a cold spot \cite{Kauppinen2015}. While IR imaging excels in detecting thermal-related defects, it is less effective in accurately identifying other types of defects that may not have thermal signatures, such as micro-cracks compared to EL or RGB imaging. IR imaging is greatly affected by ambient conditions and its quality can be compromised by solar glares and reflections \cite{Kauppinen2015}. To mitigate this issue, IR images can be cross-referenced with corresponding RGB images, enabling the identification and removal of images affected by solar reflections.

Common defects that can be identified from IR images are:

\noindent \textbf{Hot spots}. 
Hot spots are areas of localized heating which are clearly visible through the IR images of the solar panel. Hot spots can be caused by many reasons such as partial shading, cracks, or other defects in the solar cells. A recent study revealed a significant correlation between the occurrence of cracks in solar cells and the presence of hot spots, which are affected by shading ratios~\cite{Dhimish2021}.

\noindent \textbf{Open circuit and short circuit defects}.
Open circuit defect refers to a malfunction that occurs when a portion of a solar panel gets disconnected, causing the current of all the solar panels associated with the same string to be zero \cite{article}. Short circuit defect occurs when a portion of a solar panel is short-circuited, causing the current of all the solar panels associated with the same string to be higher than normal. IR imaging can detect open circuit faults in PV modules by identifying temperature increases. When a module experiences an open circuit, current flow is disrupted causing the module to heat up compared to surrounding healthy modules. This temperature rise can be detected using IR imaging, pinpointing the faulty module. IR imaging can also be used to detect short circuit faults in PV modules. A short circuit causes a localized hot spot in the affected module due to the high current flow. These hot spots can be observed in IR images, enabling the detection of short circuit faults.

\noindent \textbf{Potential-induced degradation (PID)}.
Module PID defect in solar cells refers to loss of efficiency over time caused by potential-induced degradation (PID). It occurs due to leakage of current caused by the high potential difference between the PV module and the ground. 
PID occurs due to the negative potential each PV module experiences during normal operational conditions. PID in crystalline silicon solar panels can lead to significant power loss, ranging from 30\% to 70\% \cite{solar3020019}. In IR images of PV modules, areas affected by PID appear as hot spots or regions with elevated temperatures compared to the rest of the module. These hot spots indicate the presence of increased leakage currents and can be used to identify PID-affected modules.

\subsection{Electroluminescence (EL) images}
EL is observed in solar cells when an electric current is passed through them, resulting in the conversion of electrical energy into light energy. To capture EL, a specialized camera with filters optimized for sensitivity in the near-infrared spectrum is required. EL imaging allows the visualization of current distribution in the PV module, thus aiding in defect detection. Applying a forward bias to the solar cell induces a forward current that injects numerous unbalanced carriers into the device, leading to injection EL. Given that the band gap of intrinsic silicon is approximately 1.12 eV, we can deduce that the peak wavelength of the infrared light generated through this process should be around 1150 nm. \cite{Meng}.

\begin{figure}
    \centering
    \includegraphics[width=0.5\linewidth]{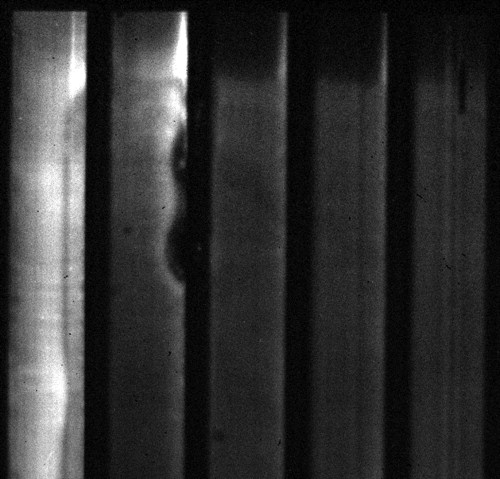}
    \caption{Electroluminescence image of an organic photovoltaic module~\cite{hoyer2010electroluminescence}. The dark region that is non-periodic indicates the presence of a defect. Reproduced with permission from AIP publishing.}
    \label{fig:eldefect}
\end{figure}

The illuminated surface in an EL image corresponds to the active area of the solar cell surface where the photovoltaic effect takes place, resulting in the generation of electric current. Darker regions appearing in a periodic manner represent the bus-bar connections used to collect and transport the generated current. Any other sporadically occurring darker regions indicate the presence of defective areas within the cell that do not contribute to power generation \cite{hoyer2010electroluminescence}. These features are demonstrated in \autoref{fig:eldefect} which is an EL image of an organic photovoltaic module.

In EL images, a monocrystalline silicon solar cell without defects exhibits a clear surface, while a polycrystalline silicon solar cell without defects shows irregular dark regions caused by crystalline grain boundaries. This complicates the detection and classification of defects on polycrystalline silicon solar cells \cite{[3]}.  EL inspection has higher cell surface resolution compared to IR imaging, enabling more accurate problem diagnosis in certain cases. Hence, EL examination is a useful method for detecting flaws like micro-cracks, grid line defects, and corroded interconnection ribbons \cite{[2]}. However, it is important to note that EL cannot quantify the impact on the module's power output. EL imaging is a non-invasive technique; however, it needs to be conducted when the farm is not in operation, typically during the night. Additionally, an external source is required to energize the cells for the imaging process. Common defects that can be detected from EL images are: 

\noindent \textbf{Microcracks}.
Microcracks are tiny fractures on solar cells that are invisible to the naked eye. They can be caused by environmental factors such as wind, hail, and heavy snowfall, or introduced during manufacturing due to poor quality or process control.  The images captured through EL imaging can allow for the early detection of microcracks. The smallest crack size that can be detected using a specific EL imaging system is in a range of 200-700 $\mu$m \cite{8733881}. The length scale of microcracks can vary, with some spanning the whole cell and others appearing in only small sections of a cell.

\noindent \textbf{Finger Interruptions}.
Finger interruptions are defects that are generated during the cell metallization and module interconnection process, often resulting in poor performance of the solar cell due to high effective series resistance \cite{8016569}. EL imaging can detect finger interruptions, which can help in identifying the specific areas of the solar panel that need further investigation and repair.

Besides these defects, EL images can also be used to detect hot spots that can have detrimental effects such as cell or glass cracking, melting of solder, or degradation of the solar cell and PID which can cause a decrease in the fill factor, short-circuit current, and open-circuit voltage, leading to worse overall performance of the solar panel.

\subsection{RGB Images}
Traditional RGB or visible spectrum imaging is a widely accessible method for detecting various solar cell defects, as it does not require specialized sensors. However, RGB imaging may not have sufficient sensitivity to detect subtle or small defects in solar cells. Some defects, such as micro-cracks or tiny impurities, may not be distinguished through RGB imaging alone \cite{rahaman2022infrared}. RGB imaging is often accompanied by extensive data preprocessing to generate usable data for defect classification technques. The effectiveness of RGB imaging for defect detection can be influenced by environmental factors, such as lighting conditions, reflections, shadows, or variations in surface texture. These factors can impact the accuracy and reliability of defect identification. RGB imaging is noninvasive and can be used to detect defects such as yellowing, grid line corrosion, snail trail, and shading~\cite{8478340}.

\noindent \textbf{Yellowing}.
Yellowing in solar panels is a phenomenon that occurs due to the degradation of ethyl vinyl acetate (EVA), a material used as an encapsulant on the panel \cite{Pern1991}. The EVA turns yellow due to UV radiation, which cannot be avoided as solar panels need to be exposed to solar radiation to produce electricity. The yellowing of PV modules is considered to reduce the efficiency of the modules, with studies showing a reduction in efficiency by 9\% to even 50\% \cite{Segbefia2023-ym}. The acetic acid released while EVA is degrading may cause corrosion in solar panels.

\noindent \textbf{Grid line corrosion}.
Grid line corrosion in solar panels is a phenomenon that occurs due to the degradation of the grid lines, typically made of silver on solar cells. This corrosion is often caused by the presence of moisture and acetic acid from the degradation of the encapsulant material such as EVA \cite{Wohlgemuth2015}.

\noindent \textbf{Snail trail}.
Snail trails are dark lines that appear on the surface of the solar panel in a spiral or snail-like pattern. Their optical effect corresponds to microcracks in the cells, consisting of discoloration in the paste used in the silkscreen of the photovoltaic cells~\cite{Liu2015}.

\noindent \textbf{Shading defect}.
Shading defect in solar panels occurs when a part of the panel gets shaded and does not contribute to power generation. This can significantly lower the performance of the panel \cite{Abdelaziz2022}.

\section{Method of data collection}

The method of data collection is often an overlooked aspect of defect detection, yet it plays a crucial role in influencing the outcome of trained models. The standard data collection approach for inspecting solar fields begins with the mounting of a sensor on a UAV, which is then deployed to fly above the solar field. During the flight, the sensor gathers the data, for example, IR images and other relevant information such as the location of the image or the flight path, which is subsequently transmitted wirelessly to a ground control station. Upon reaching the ground control station, the collected images and data are stored in  a dedicated data storage facility for further processing. This stage often involves the organization, categorization, and initial analysis of the collected data. Various data preprocessing techniques are applied to improve the quality of the collected data. A machine learning algorithm is used at the last step to extract valuable insights and detect any potential defects or irregularities within the solar field.
 While automated systems can improve efficiency, there are limitations to current sensing technologies. No single technique is adequate for all conditions, which can lead to missed defects or false positives during inspections \cite{mahmoud_meribout__2023}. We summarize two representative approaches used for data collection and highlight the key differences.

\begin{figure}
    \centering
    \includegraphics[width=\textwidth]{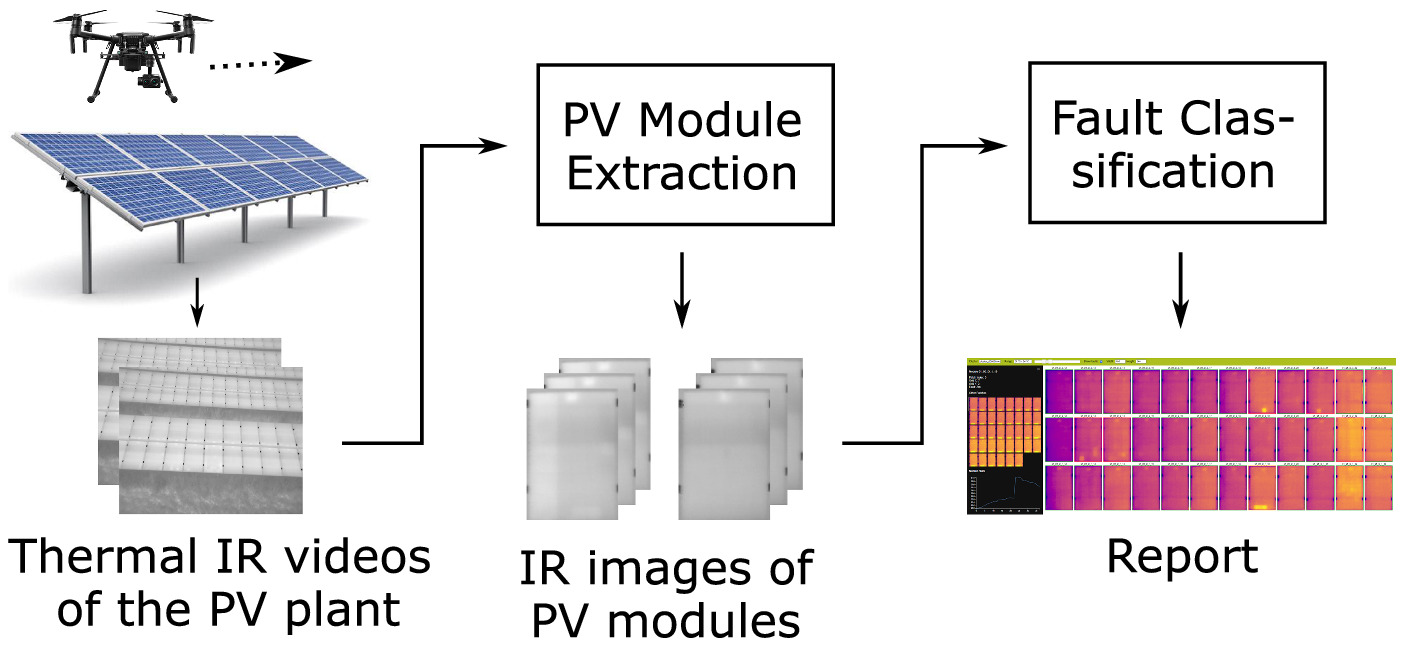}
    \caption{Overview of defect detection tool using IR images developed by Bommes et al.~\cite{https://doi.org/10.1002/pip.3448}. The data collection is done through aerial videos recorded by UAV.}
    \label{fig:bommes}
\end{figure}

Bommes et al.~\cite{https://doi.org/10.1002/pip.3448} present a comprehensive and robust image data collection system for solar panels as shown in \autoref{fig:bommes}.  The image collection system utilizes UAV with IR video recording, as opposed to the traditional method of collecting images. This approach enables the capture of multiple pictures of each panel, facilitating the elimination of images containing noise and disturbances. Consequently, it allows for the selection of the best images, resulting in a substantial dataset comprising 4.3 million frames. The path followed by the UAV is critical as it determines the labeling of the solar panels.
This procedure requires significant preprocessing of the videos including the 
 extraction of modules from the frames using mask R-CNN (Region-based convolutional neural network) instance segmentation. 

An alternative approach was adopted in another study~\cite{NAVEENVENKATESH2022110786} that generated and used only 600 RGB images, with 100 images for each defect type. Unlike the previous study, this work collected images under ideal laboratory conditions using UAVs, ensuring optimal image quality without any compromises. However, this might limit the model's robustness when applied to real-world datasets, where image quality may be affected by factors such as UAV movement in windy conditions, solar glare, and other environmental disturbances. This highlights the importance of considering real-world challenges when developing and training defect detection models. A generative adversarial network (GAN) was employed to expand the dataset to 3150 images. These two approaches demonstrate diverse data collection and augmentation methodologies, each with its unique set of advantages and challenges.

\section{Data preprocessing}

Preprocessing of image data obtained from UAVs is essential to prepare it for use as input in machine learning models. Furthermore, preprocessing enhances data quality and quantity through techniques such as data augmentation, noise reduction, and artifact removal, eventually leading to improvement in the performance of the defect classification algorithm.

\subsection{Image processing techniques}
Grayscale conversion is a common preprocessing technique for IR and EL images in defect detection, as it facilitates data standardization and enhances model performance. This is because temperature differentials, rather than color, are the primary focus. Grayscale images effectively represent these temperature variations, aiding in the identification of potential defects based on temperature differences. Additionally, grayscale images have smaller file sizes, making them easier to store, transmit, and process in large-scale data collection scenarios like solar field inspection.

When using RGB images for defect detection, minimal preprocessing such as filtering or grayscale conversion is preferred to preserve the color and texture information vital for identifying visual defects on the solar panels. However, certain enhancement processes can be applied to RGB images to improve their quality and clarity. These processes include Gaussian blurring to reduce unwanted visual artifacts caused by noise \cite{gedraite2011investigation},  unsharp masking or high-pass filtering to enhance edge and detail definition within the image, and histogram equalization or contrast stretching to improve the visual distinction between different elements in the image.

Ref. \cite{https://doi.org/10.1002/pip.3448} discusses a detailed methodology for preprocessing before defect detection. The author uses a mask-RCNN to perform segmentation of the solar panels, starting with MS COCO pre-trained weights~\cite{10.1007/978-3-319-10602-1_48} followed by training on the generated dataset. The mask-RCNN model achieves an F1 score of 98.92\% and an average precision of 99.55\% at an IoU (intersection over union score) threshold of 0.5. The main issues during data extraction involved row filtering failure, panel segmentation failure, and irregular drone flight paths. The tool fails only for 49 out of 561 PV plant rows, which corresponds to only 12.2\% of all PV modules.
To tackle the problem of solar glares on the extracted PV modules, the authors demonstrate an effective use of filters to remove patches affected by non-stationary reflection-type solar glares, preventing issues in downstream anomaly classification. These filters were also used to identify the rows of solar panels during preprocessing. Furthermore, filters can be employed to isolate individual solar panel images from an array.

The approach presented in Ref. \cite{Patel} showcases the potential of traditional data processing in detecting damaged areas in solar panels. By leveraging filters and image processing techniques, such as morphological erosion and blob detection, damaged areas were successfully identified, and bounding boxes were determined. These techniques can enhance data quality during the collection process. For the segmentation of individual components in objects, an architecture consisting of mask R-CNN followed by Bayesian fully convolutional networks  has proven to be  effective \cite{bayesianNN}. Such Bayesian networks may find applications in image processing for PV modules.

\subsection{Data augmentation}
Data augmentation is crucial for overcoming the challenges posed by limited annotated data. By generating augmented images, the model develops greater resilience to variations in lighting conditions, solar panel orientations, and defect types. This results in a more generalized and effective defect detection model, capable of accurately identifying defects across diverse scenarios.

CNN models exhibit translational equivariance, whereby the network's output remains consistent under spatial translations. However, they lack rotational equivariance, implying that the feature vectors do not consistently rotate in a predictable manner when the input undergoes rotation \cite{harmonic-net-equivariance}. Therefore, data augmentation procedures that involve rotation and scale operations are frequently used in preprocessing \cite{NAVEENVENKATESH2022110786}. Moreover, augmentations involving intensity, brightness, and contrast of the data could lead to better generalization under ambient conditions.

\begin{figure}[htbp]
    \centering
    \includegraphics[width=15cm]{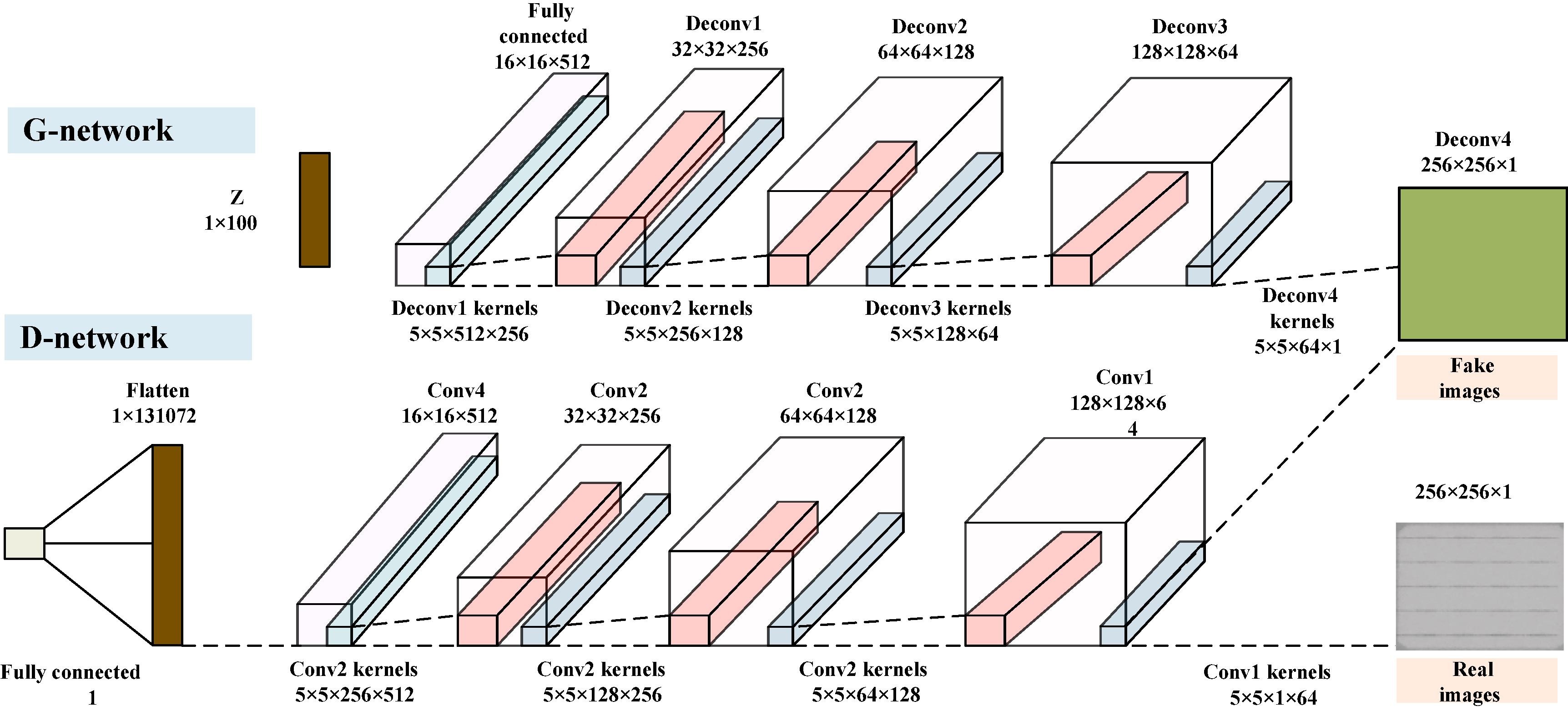}
    \caption{%
    Structure of the GANs used for data augmentation in Ref. \cite{TANG2020453}
    The G-network (generator) creates synthetic images from random noise or a latent vector to mimic the characteristics of the original images. Its goal is to make these images indistinguishable from real ones. Meanwhile, the D-network (discriminator) acts as a classifier, distinguishing between images generated by the G-network and the original ones.  \autoref{fig:GAN generated images} showcases the images generated by the G-network, demonstrating the effectiveness of the GAN in producing realistic visual outputs. Reproduced with permission from Elsevier.}
    \label{fig:GAN}
\end{figure}

\begin{figure}[htbp]
    \centering
    \includegraphics[width=15cm]{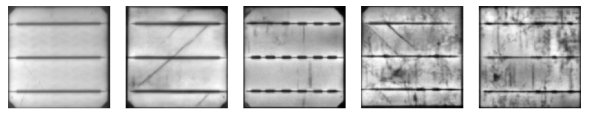}
    \caption{Images of solar cell generated using GAN in Ref. \cite{TANG2020453}. Reproduced with permission from Elsevier.}
    \label{fig:GAN generated images}
\end{figure}

Generative adversarial networks (GANs) are gaining popularity for data augmentation as they can effectively extract deep image features to facilitate high-quality augmentation. The traditional architecture of GANs with two networks, generative  G-network and D-network is illustrated in \autoref{fig:GAN}. The G-network is responsible for generating synthetic images that mimic the characteristics of the original image while the D-network acts as a classifier to distinguish them from the original images. 
Tang et al.~\cite{TANG2020453} used a combination of GANs and traditional argumentation to generate the data, increasing the dataset size from 1800 to 8600 images. The approach uses Wasserstein GAN (WGAN) as it outperformed deep convolutional GAN (DCGAN) in data argumentation of EL images.
The augmented images shown in \autoref{fig:GAN generated images} exhibit significant variations while introducing randomness, helping prevent overfitting~\cite{TANG2020453}. %
However, a major challenge to the adoption of GANs for data augmentation is the lack of an intrinsic metric for accurately quantifying and evaluating its performance \cite{devries2019evaluation}. While some experimental metrics have been proposed such as  Fréchet Joint Distance (FJD) \cite{devries2019evaluation}, perceptual similarity metrics such as LPIPS (Learned Perceptual Image Patch Similarity) \cite{barua2019qualityevaluationgansusing} and Inception Score (IS), further investigation in this direction is warranted. Additionally, the training of GANs is often unstable and requires substantial computational resources \cite{gulrajani2017improved}. 

\section{Machine learning approaches for defect detection}

\begin{table}[htbp]
\caption{Overview of some defect detection approaches summarized in this review.}
\label{tab:papersumm}
\footnotesize

\begin{tabular}{p{2.5cm} p{2cm} p{3cm} p{3cm} p{2cm} p{2cm}}
\hline
\textbf{Reference} & \textbf{Image type} & \textbf{Source} & \textbf{Dataset size} & \textbf{Accuracy} & \textbf{Architecture}\\
\hline

Lawrence Pratt et al. \cite{[3]} & EL images & Combination of public and private sources & 148 images & - & UNET \\
\hline

Xiaoxia Li et al. \cite{8478340} & RGB images & Custom dataset & 8400 images & 98.5\% & CNN \\
\hline

Bommes et al. \cite{https://doi.org/10.1002/pip.3448} & IR images & Custom dataset & 4.3 million (frames extracted from video) & 90.91\% & Mask RCNN - ResNet-50 \\
\hline

Wuqin Tang et al. \cite{TANG2020453} & EL images & Public and private sources & 8600 (using GAN); 1800 originally & 83\% & GAN - Custom CNN\\
\hline

Naveen Venkatesh et al. \cite{NAVEENVENKATESH2022110786} & RGB images & Lab-based images collection & 3150 (after augmentation); 600 originally & 98.95\% & AlexNet - J48 decision tree - Lazy classifiers\\
\hline

Sergiu Deitsch et al. \cite{Deitsch_2019} & EL images & Custom dataset & 2624 images & CNN (88.42\%), SVM (82.44\%) & CNN, SVM\\
\hline

Imad Zyou et al. \cite{inproceedings} & RGB images & Custom dataset & 599 images & 93.3\% & AlexNet \\
\hline

Binyi Su et al. \cite{9398560} & EL images & Custom dataset & 3629 images & Cracks (73.16\%), fingerprint (91.3\%), black core (100\%) & BAFPN\\
\hline

Chen et al. \cite{Chen2020-fc} & RGB images & Custom Dataset & around 24000 images & 93.23\% & Combination of random forest with CNN\\
\hline

Alexander Bartler et al. \cite{bartlerAutomatedDetectionSolar2018} & EL images & Custom Dataset & 98,280 labeled cell images & (BER) 7.73\% & CNN\\
\hline
\end{tabular}
\end{table}

The choice of the machine learning algorithm is arguably the most important factor determining the accuracy of defect detection and classification. Some representative machine learning approaches for defect detection are summarized in \autoref{tab:papersumm}. Overall, the machine learning approaches used can be broadly classified into three types: traditional machine learning, e.g., decision trees, support vector machine (SVM); deep learning approaches involving neural network architectures; and hybrid approaches that combine the two.

The advent of deep learning and accelerated training capabilities using graphics processing units (GPUs) has led to a gradual decline in the use of traditional machine learning approaches. As deep learning models continue to demonstrate superior performance and efficiency, they are increasingly becoming the preferred choice for many applications.
This trend was observed in the reviewed articles, as most of the approaches used some form of deep learning. These models employing CNNs acting on image data can automatically learn hierarchical representations of data \cite{Bilal_2018}. They are capable of handling large amounts of data and capturing complex patterns. Additionally, deep learning approaches can adapt to new features and patterns in the data without manual feature engineering, a common requirement in traditional machine learning algorithms. This flexibility often results in higher accuracy of defect classification.  This is evident from the study by Deitsch et al. \cite{Deitsch_2019}, where both CNN and SVM models were trained to perform defect detection on EL images of solar modules.  Though this study didn't distinguish defect types, the CNN (88.42\% accuracy) outperforms the SVM model (82.44\%) in average classification accuracy. We now discuss the deep learning approaches used in literature for defect classification.

\begin{figure}
   \centering
    \includegraphics[width=0.8\textwidth]{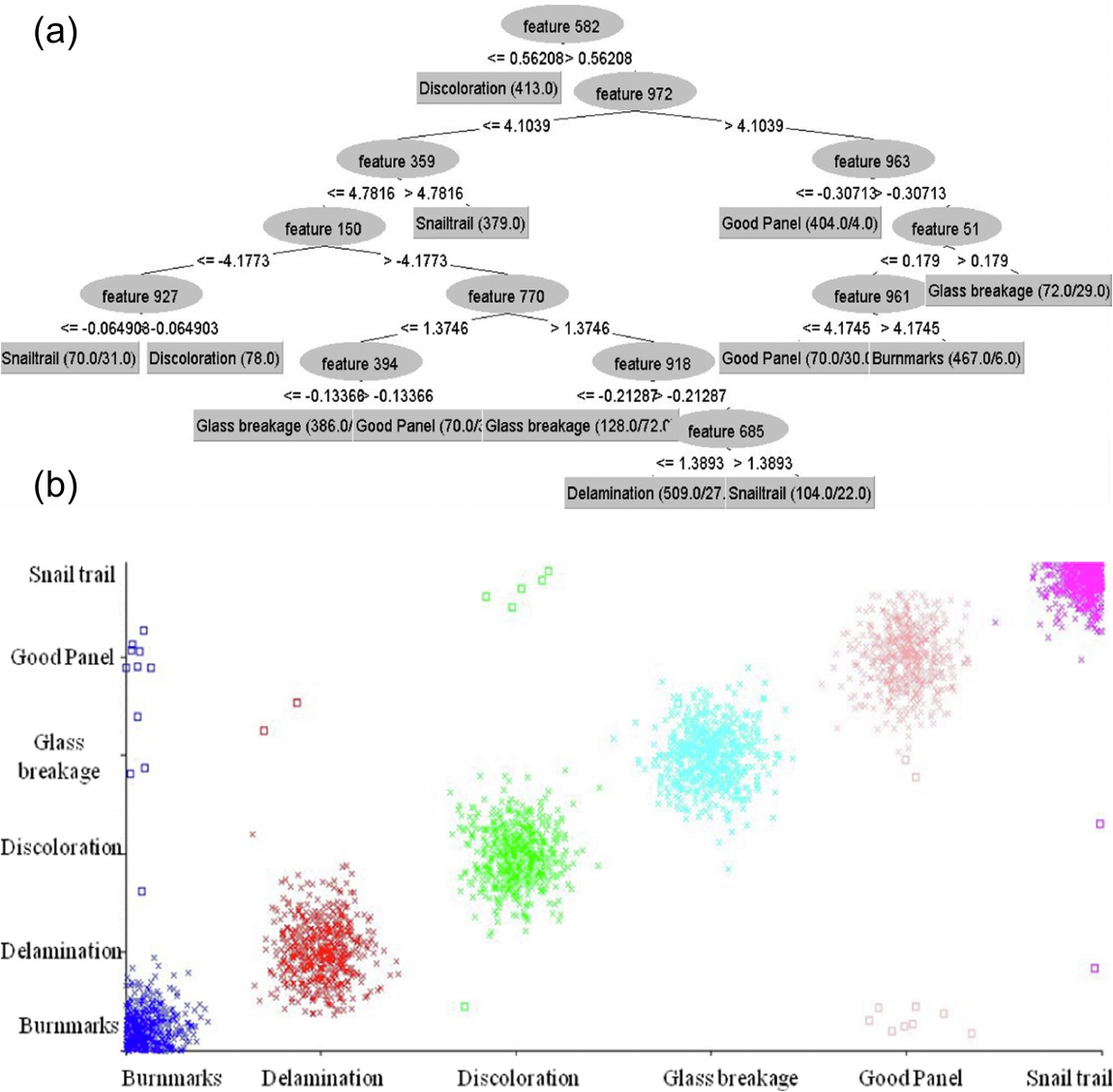}
    \caption{ Defect detection in PV modules using hybrid approach \cite{NAVEENVENKATESH2022110786}. Deep learning is used to extract features from images which are then used in a decision tree algorithm.
    (a) J-48 decision tree algorithm used defect classification. From the structure of the decision tree, we can interpret the criteria used for the classification. 
    (b) Results of defect classification using the k-nearest neighbor (KNN) algorithm. The KNN algorithm effectively separates the defect classes into different regions. Reproduced with permission from Elsevier.
    }
    \label{knn-j48}
\end{figure}

Li et al. \cite{8478340} present a custom CNN designed to identify defects in RGB images of solar panels. The defects include dust shading, encapsulant delamination, gridline corrosion, snail trails, and yellowing. 
The proposed solution achieves an impressive accuracy of 98.7\% for defect classification. The authors also explore the interpretability of the model by generating t-SNE plots (t-distributed stochastic neighbor embedding) to analyze the deep features extracted from various layers of the CNN. The t-SNE analysis provides insights into the performance differences between various models, as better-performing models show a clear separation of defect classes in the t-SNE maps. Additionally, the t-SNE maps become more distinguishable for features extracted from deeper layers.

Bartler et al. \cite{bartlerAutomatedDetectionSolar2018} employed a standard CNN architecture for automated detection in solar cells. Their dataset consisted of 98,280 labeled EL  images of cells, derived from 1,366 module images. Oversampling and data augmentation techniques were used to address significant data imbalance. Instead of multi-class classification, their model performed binary classification, distinguishing between normal and defective solar cells. This approach achieved a bit error rate  of 7.73\% for binary classification in real-world experiments.

\begin{figure}
    \centering
    \includegraphics[width=\textwidth]{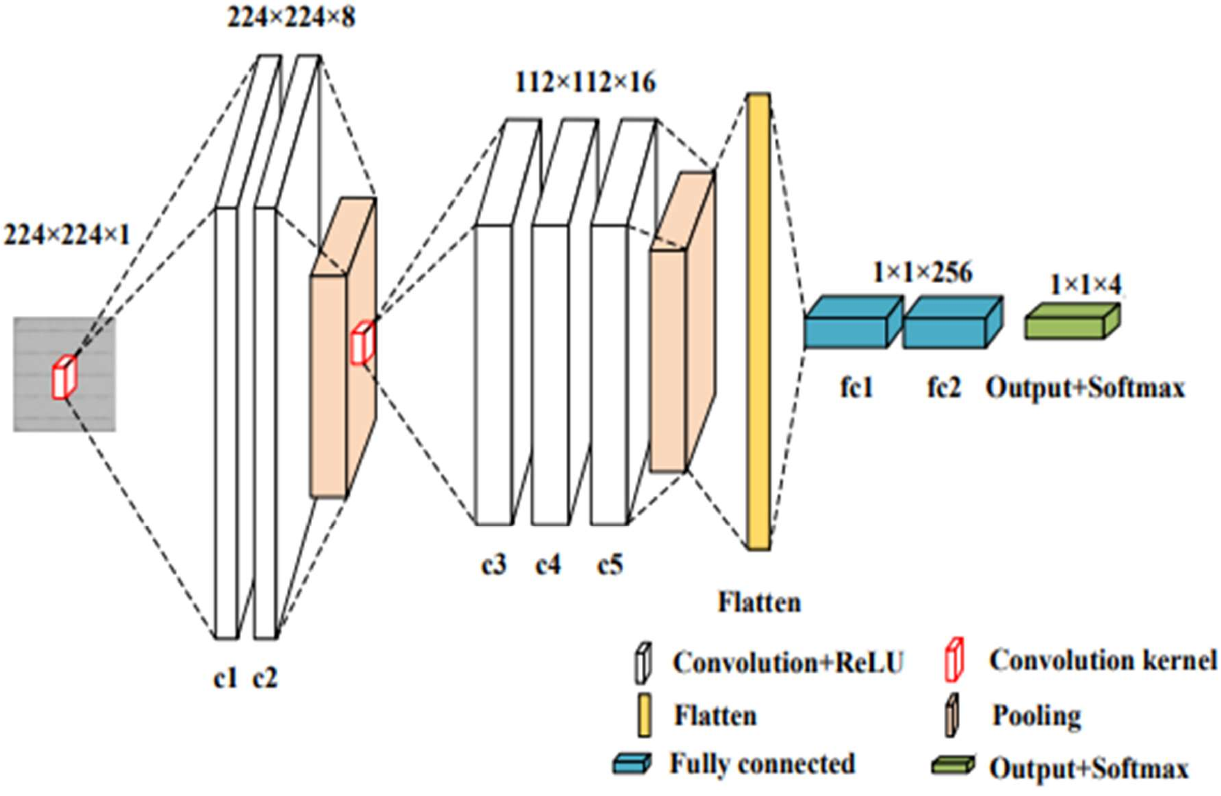 }
    \caption{Standard CNN architecture  used in Ref. \cite{TANG2020453}. The architecture used a total of 5 convolutional layers and 2 pooling layers followed by a standard dense layer with softmax function for classification.
    Reproduced with permission from Elsevier.}
    \label{fig:CNN in TANG}
\end{figure}

Tang et al. \cite{TANG2020453} performed defect detection on both polycrystalline-Si and monocrystalline-Si using EL images. The defects classified were micro-cracks, finger interruptions, and breaks. A key feature of their approach was the use of a combination of  GANs (\autoref{fig:GAN}) and traditional data augmentation for increasing the dataset size. The proposed augmentation approach was compared with approaches involving only GANs or only traditional data augmentation, with the latter two methods achieving only 74\% and 56\% accuracy, respectively. In contrast, the custom CNN architecture proposed in the study \autoref{fig:CNN in TANG} achieves an accuracy of 83\%. The authors also compare the complexity of the model with other approaches, such as transfer learning, based on the total number of model parameters and trainable parameters. These parameters are the ones that are updated during the training process to minimize the loss function, while non-trainable parameters are fixed and do not change. Therefore, the trainable parameters directly impact the model's ability to learn from the data and make accurate predictions. Their analysis revealed that their model provides a good tradeoff between accuracy and model complexity.

Venkatesh et al. \cite{NAVEENVENKATESH2022110786} implemented a hybrid approach for defect detection combining deep learning and traditional machine learning.  A pre-trained AlexNet model was used to extract 1000 features from each image of the solar cell. 
Subsequently, a J48 decision tree algorithm as shown in \autoref{knn-j48}a was utilized for feature selection and dimensionality reduction. Finally, a series of lazy learning algorithms were compared for classification accuracy. The k-nearest neighbor (KNN) algorithm achieved a high accuracy of 98.95\% and a quick build time of only 0.04 seconds, the results of which are shown in \autoref{knn-j48}b.

Bommes et al. \cite{https://doi.org/10.1002/pip.3448} used a ResNet-50 model to perform classification of ten common thermal anomalies, achieving an accuracy of 89.40\%. Their data collection approach produced multiple images of a single module, thereby allowing multiple classifications. The multiple classifications when combined using majority voting, boosted the accuracy to 90.91\%. The low accuracy on certain defect types was associated with the under-representation of those defect classes in the dataset. These issues could be solved using GANs to produce artificially generated images of those classes, leading to a more balanced dataset.

\begin{figure}
    \centering
    \includegraphics[width=\textwidth]{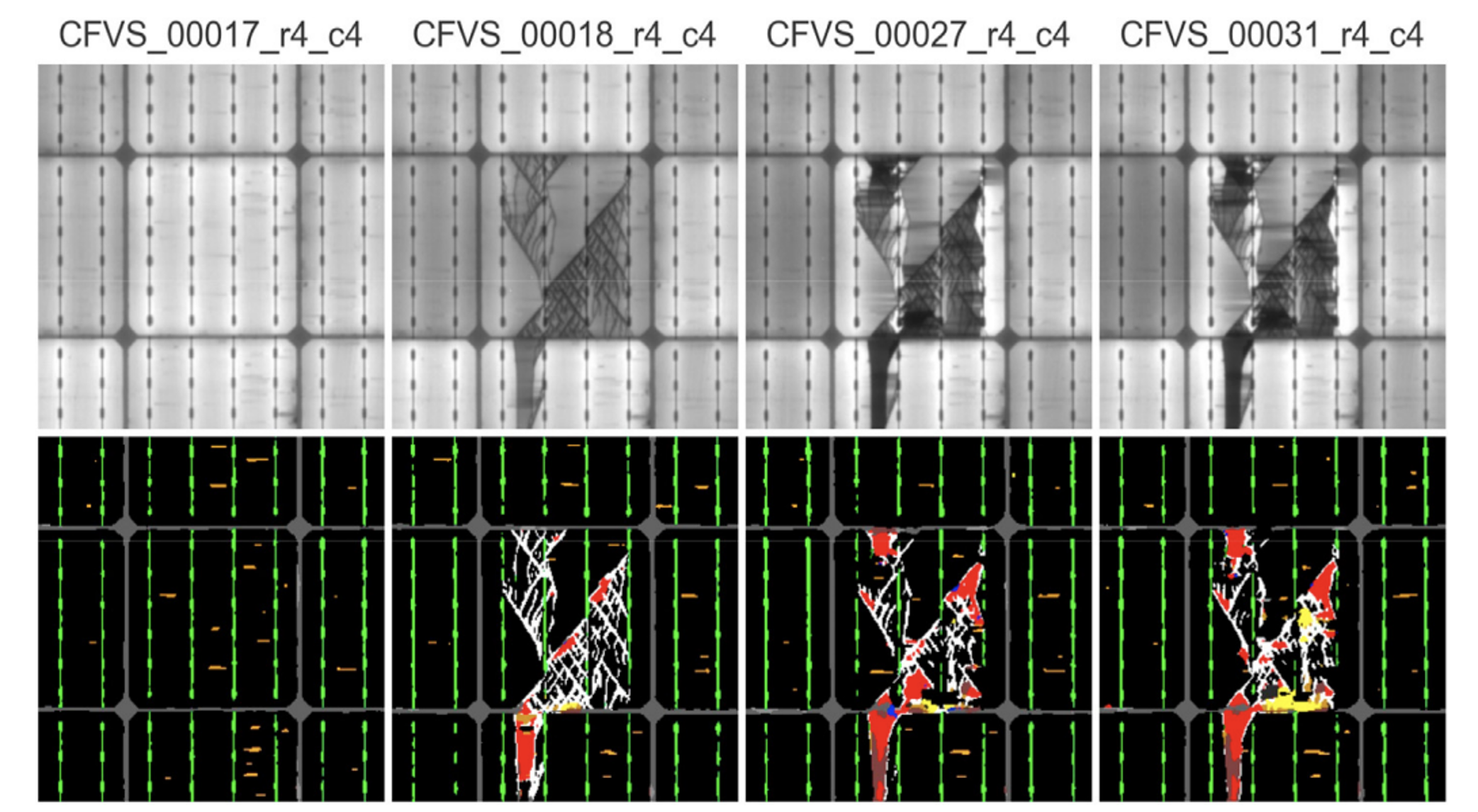}
    \caption{The output of semantic segmentation, as demonstrated in Ref. \cite{[3]}, highlights the impressive performance of the model in identifying breaks and cracks within solar cells. The top row displays the EL images, while the bottom row displays the output after the image segmentation has been completed. The semantic segmentation model excels in this task by accurately delineating and categorizing regions of interest within images, specifically isolating areas indicative of structural damage.
    Identifying this damage is crucial for maintenance and quality assessment purposes. Reproduced with permission from Elsevier.}
    \label{fig:Semanticsegmentation}
\end{figure}

Pratt et al. \cite{[3],PRATT2023200048} developed a semantic segmentation model based on U-NET architecture for defect classification. The unique aspect of this model is that the semantic segmentation approach generates pixel-level classification about defects in the solar cell. This information is helpful especially when multiple defects are present in the same cell in different areas as the model can predict the probability of each defect in a solar cell.  The segmented images alongside the EL images are  shown in \autoref{fig:Semanticsegmentation}. This method used module-level images of solar panels for training. Initially, the EL image of the complete solar panel was taken, from which module-level images were extracted. These module-level images were cropped such that each image contained images of  half-cells surrounding the central full cell. This allows the model to detect whether the cell is made of monocrystalline or polycrystalline silicon because monocrystalline  cells have distinct cut-outs from the corners while polycrystalline cells have regular corners. This buffer provided by the half-cell images around the main cell guarantees that the cell in the center is completely visible and does not exhibit edge trimming during cropping. 
The model was trained to distinguish between intrinsic features of the module such as bus bars, ribbon connection, and module borders which may be visible in the EL image, as well as extrinsic defects. The approach utilizes a pre-trained VGG-16 as the base model (encoder) and U-NET as the semantic segmentation model (decoder) ~\cite{guptasegmentation2020}.  The model performs well for defect classification on both monocrystalline and polycrystalline silicon while detecting the intrinsic features. The performance was significantly better on the monocrystalline modules during defect detection which was attributed to the larger training sample size and smoother module surfaces without grain boundaries.

Su et al. \cite{9398560} introduced the bidirectional attention feature pyramid network (BAFPN), a novel approach for pinpointing defect locations on solar panels. They addressed a common issue in deep learning models, particularly CNNs, where increasing network depth can lead to feature vanishing, making small defect detection challenging. BAFPN captures refined information across different scales, ensuring the retention of small defect features.
The BAF-Detector, which integrates BAFPN into Faster RCNN+FPN, achieves robust multi-scale defect detection performance. This is attributed to the multi-scale feature fusion, which enables all pyramid layers to share similar semantic features. This improves the network's robustness to scales and enhances the detection of small defects. Additionally, the attention mechanism using cosine similarity helps to highlight defect features while suppressing complex background features, leading to more accurate defect detection.
Experimental results on a large-scale dataset demonstrate BAF-detector's effectiveness, yielding exception performance for classification metrics,  F-measure, mAP, and IoU scores in classifying and detecting defects in raw PV cell EL images.

\begin{figure}[h]
    \centering
    \includegraphics[width=\textwidth]{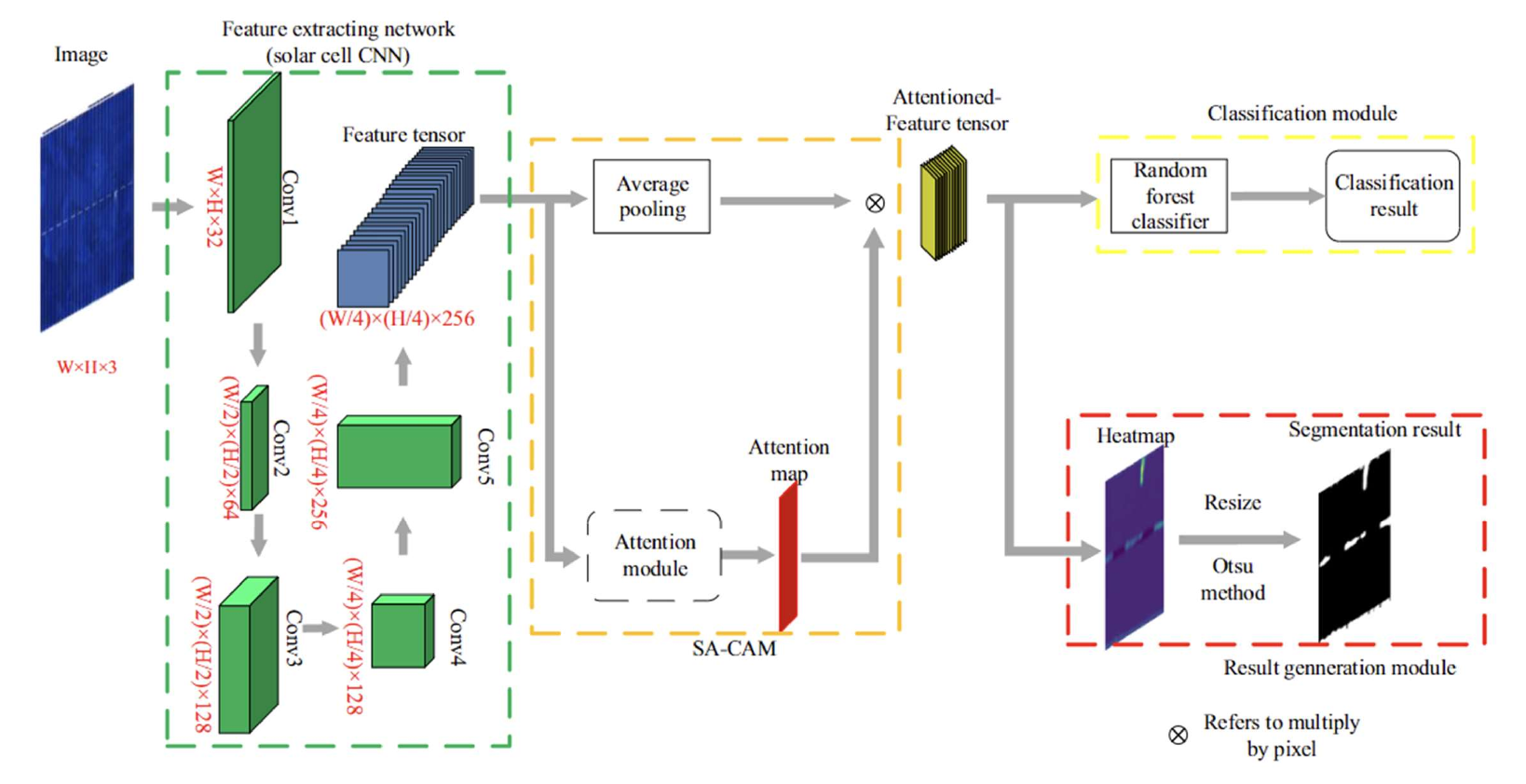}
    \caption{Robust weakly supervised learning of deep ConvNets for surface defect inspection from Ref. \cite{Chen2020-fc}.
    In this approach, deep ConvNets are used to automatically learn and extract hierarchical features from images, particularly focusing on surfaces such as those found in industrial settings. The image is first passed to the feature-extracting network, which utilizes CNNs to extract the feature tensor from the given image. This feature tensor is then passed onto the SA-CAM (Semantic Attention Class Activation Map) block, which is used to localize surface defects in images without requiring pixel-level annotations during training. The generated attentioned-feature tensor can be then used to perform either classification tasks or can also be used to generate heat maps to generate image segmentation results.  
    }
    \label{fig:RWSLDC}
\end{figure}

Chen et al. \cite{Chen2020-fc}  propose a novel end-to-end framework for defect classification and segmentation leveraging weakly supervised learning of a CNN with attention architecture as shown in \autoref{fig:RWSLDC}. The framework integrates a robust classifier and a spatial attention module to enhance defect feature representation and improve classification accuracy. 
A key innovation is the introduction of a spatial attention class activation map (SA-CAM), which generates precise heatmaps, suppresses background interference, and highlights defective areas effectively
The proposed framework is trained using only global image labels and is robust to complex backgrounds. The authors discuss the incorporation a random forest classifier in the CNN, which shows strong robustness and adaptability compared to traditional methods. Experimental results demonstrate the generalization of the proposed method on three distinct datasets with different textures and backgrounds, showing improvements in classification accuracy by 0.66-25.50\% and segmentation accuracy by 5.49-7.07\%

Most of the methods discussed involve transfer learning or custom CNN architecture for classifying defects in images of solar modules. When comparing pre-trained transfer learning models and custom CNN, two key factors to consider are the practicality of the approach and their performance with the size and quality of the training data.
Transfer learning models can be more practical when the dataset size is limited, as they can leverage pre-trained models on large datasets to improve performance on smaller datasets. However, traditional CNNs are the optimal choice when the dataset is large and diverse, as they can be trained from scratch to fit the specific dataset \cite{inproceedings}. CNNs can be task-specific, i.e., capturing specific details in the dataset provided there is a sufficiently large training dataset. Some studies have proposed novel architectures, such as hybrid fuzzy CNN (HFCNN) \cite{9301240} to improve the accuracy of defect detection in solar panels.
Another distinction in the approaches is the spatial resolution of the information obtained. Some models only identify modules with defects while others predict probabilities for the presence of defects in different locations within individual modules or cells. The latter are also able to identify multiple defects within a single cell or module.

\section{Interpretability of defect detection approaches}
Deep learning models, specifically CNNs, are often referred to as ``black boxes" due to their opaque and unexplainable nature \cite{liang2021explaining} compared to traditional machine learning approaches. This term reflects the inherent difficulty in understanding and interpreting the internal workings of these complex models. The complexity lies in the convolutional layers and their hierarchical feature learning, number of parameters on the order of millions, non-linear activation functions, and high-dimensional processing of data. The complexity of modern neural network architectures, characterized by multiple branches, skip connections, and attention mechanisms, significantly contributes to the opacity of their decision-making processes. The absence of intuitive human analogies, coupled with the lack of transparency during the training phase, further exacerbates this issue. The limited understanding of the underlying reasoning behind the models' predictions obstructs the adoption of rational techniques for improvement.

The interpretability of deep learning models is particularly crucial in high-stakes domains like the energy generation sector. To ensure a continued supply of energy, it is essential for solar plants to correctly estimate power generation which is dependent on efficiency and hence, the presence of defects in solar cells.  It is, therefore, imperative to determine the machine learning model's accuracy in detecting defects and the extent to which it emphasizes background information. To uncover the decision-making processes of these models, we perform  an interpretability analysis using three techniques: Local Interpretable Model-agnostic Explanations (LIME), SHapley Additive exPlanations (SHAP), and Gradient-weighted Class Activation Mapping 
 (Grad-CAM). 
To facilitate a comparison between three interpretability methods, we first applied them to a CNN model used for a task simpler than defect detection, i.e. classifying the solar module into monocrystalline-Si and polycrystalline-Si in the ELPV dataset \cite{PRATT2023200048,Buerhop2018,Deitsch2021,Deitsch_2019}.
The architecture of the model we used is shown in \autoref{tab:neural-network_image classification}. To ensure a fair comparison, we maintained consistent data preprocessing techniques and model training protocols across all three approaches. The CNN model achieved a perfect accuracy of 100\% in classifying the ELPV dataset, as shown in \autoref{tab:cnn-performance}. This is due to the relatively straightforward nature of the classification task. The outstanding performance suggests that the model identifies the distinctive characteristics of solar modules, particularly the corner features of the images.  Notably, the model has likely discerned that polycrystalline silicon solar cells exhibit a complete square shape, whereas monocrystalline silicon modules have a cut-out at each corner (see \autoref{fig:LIME_Classifying}).

\begin{table}
    \centering
    \caption{Architecture of the representative CNN, a simple network of alternative conventional layers and pooling layers whose interpretability is analyzed.}
    \label{tab:neural-network_image classification}
    \begin{tabular}{|c|c|c|c|}
        \hline
        \textbf{Layer (type)} & \textbf{Output Shape} & \textbf{Param \#} & \textbf{Details} \\
        \hline
        \texttt{conv2d} & (None, 120, 120, 16) & 448 & Conv2D \\
        \hline
        \texttt{max\_pooling2d} & (None, 60, 60, 16) & 0 & MaxPooling2D \\
        \hline
        \texttt{conv2d\_1} & (None, 60, 60, 16) & 2320 & Conv2D \\
        \hline
        \texttt{max\_pooling2d\_1} & (None, 30, 30, 16) & 0 & MaxPooling2D \\
        \hline
        \texttt{conv2d\_2} & (None, 30, 30, 32) & 4640 & Conv2D \\
        \hline
        \texttt{max\_pooling2d\_2} & (None, 15, 15, 32) & 0 & MaxPooling2D \\
        \hline
        \texttt{flatten} & (None, 7200) & 0 & Flatten \\
        \hline
        \texttt{dense} & (None, 16) & 115216 & Dense \\
        \hline
        \texttt{dropout} & (None, 16) & 0 & Dropout \\
        \hline
        \texttt{dense\_1} & (None, 1) & 17 & Dense \\
        \hline
    \end{tabular}
    \smallskip
    \begin{flushleft}
        \textbf{Total params:} 122641 (479.07 KB) \\
        \textbf{Trainable params:} 122641 (479.07 KB) \\
        \textbf{Non-trainable params:} 0 (0.00 Byte)
    \end{flushleft}
\end{table}

\begin{table}
    \centering
    \caption{The performance evaluation of a CNN (Convolutional Neural Network) in edge shape detection, as described, indicates that a simple model achieved high accuracy in this specific task. Edge shape detection typically involves identifying boundaries and contours within an image, a fundamental task in image processing and computer vision.} 
    \label{tab:cnn-performance}
    \begin{tabular}{|l|cccc|}
        \hline
        \textbf{Class} & \textbf{Precision} & \textbf{Recall} & \textbf{F1-Score} & \textbf{Support} \\
        \hline
        \texttt{monocrystalline} & 1.00 & 1.00 & 1.00 & 216 \\
        \texttt{polycrystalline} & 1.00 & 1.00 & 1.00 & 309 \\
        \hline
        \textbf{Accuracy} & \multicolumn{4}{c|}{1.00 (525)} \\
        \hline
        \textbf{Macro Avg} & 1.00 & 1.00 & 1.00 & 525 \\
        \textbf{Weighted Avg} & 1.00 & 1.00 & 1.00 & 525 \\
        \hline
    \end{tabular}
\end{table}

\subsection{Local Interpretable Model-agnostic Explanations (LIME)}
LIME was developed  in 2016 and falls under the umbrella of explainable artificial intelligence \cite{ribeiro2016why}. It is widely used to interpret the outcomes generated by machine learning models, particularly for image classification tasks. LIME aims to highlight the specific regions within an image that contribute the most to the classification decision, referred to as LIME mask.

The functioning of LIME involves the generation of synthetic data in the vicinity of the original data point of interest. Subsequently, a simpler and more interpretable model is trained on this synthetic data, which is employed to explain the predictions made by the original, more complex model. 
The weights associated with the simpler model are then used to explain the impact of each feature on the original model's prediction. LIME can explain the predictions of any machine learning model, irrespective of its complexity or the nature of the data.

\begin{figure}
    \centering
    \includegraphics[width=\textwidth]{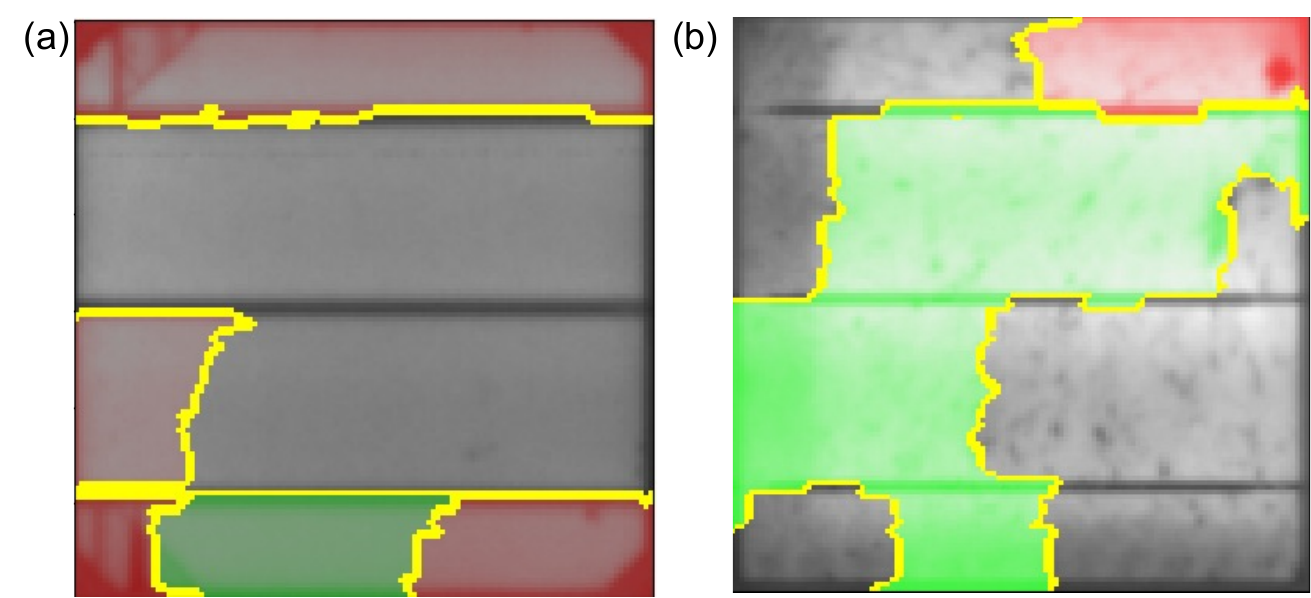}
    \caption{LIME mask for (a) monocrystalline and (b) polycrystalline silicon. The LIME heatmap overlaid on top of images shows prominent red-colored areas concentrated towards specific corners or regions. These areas strongly emphasize the features or patterns the model relies on to classify the silicon modules. While the model emphasizes corner regions in monocrystalline silicon, no such region can be identified for polycrystalline silicon.}
    \label{fig:LIME_Classifying}
\end{figure}

We applied the LIME method to a CNN model that perfectly classifies EL images of solar modules obtained from the ELPV dataset into polycrystalline and monocrystalline silicon classes. 
\autoref{fig:LIME_Classifying} shows the LIME mask for monocrystalline and polycrystalline silicon. It is evident that the model focuses on the corners of the silicon modules. However, the LIME mask generated for polycrystalline silicon modules does not reveal the same behavior. 
While the LIME method is straightforward to implement, there are instances where the generated masks may lack clear interpretability regarding the model's focus. In such cases, multiple masks may be needed to discern the underlying patterns captured by the model.

\subsection{Shapley Additive Explanations (SHAP)}
SHAP is based on the concept of Shapley values from cooperative game theory \cite{faigle1992shapley}, which are used to distribute gains to each player fairly. SHAP measures the contribution of each feature in the prediction outcome. Across all possible feature combinations, SHAP values provide a fair and consistent way to attribute the importance of features in a prediction, allowing for better understanding and trust in the model's predictions. Mitra et al. \cite{mitra2023novel} compared the explanations provided by LIME and SHAP by developing a distance metric. They found that the average difference between the two techniques varies significantly depending on the type of learning task. The differences may arise due to the inherent properties of the explainers and learning tasks. Compared to LIME, SHAP can be computationally expensive, especially for large datasets and complex models \cite{ledel2022studying}.

\begin{figure}[ht]
    \centering
    \includegraphics[width=0.8\textwidth]{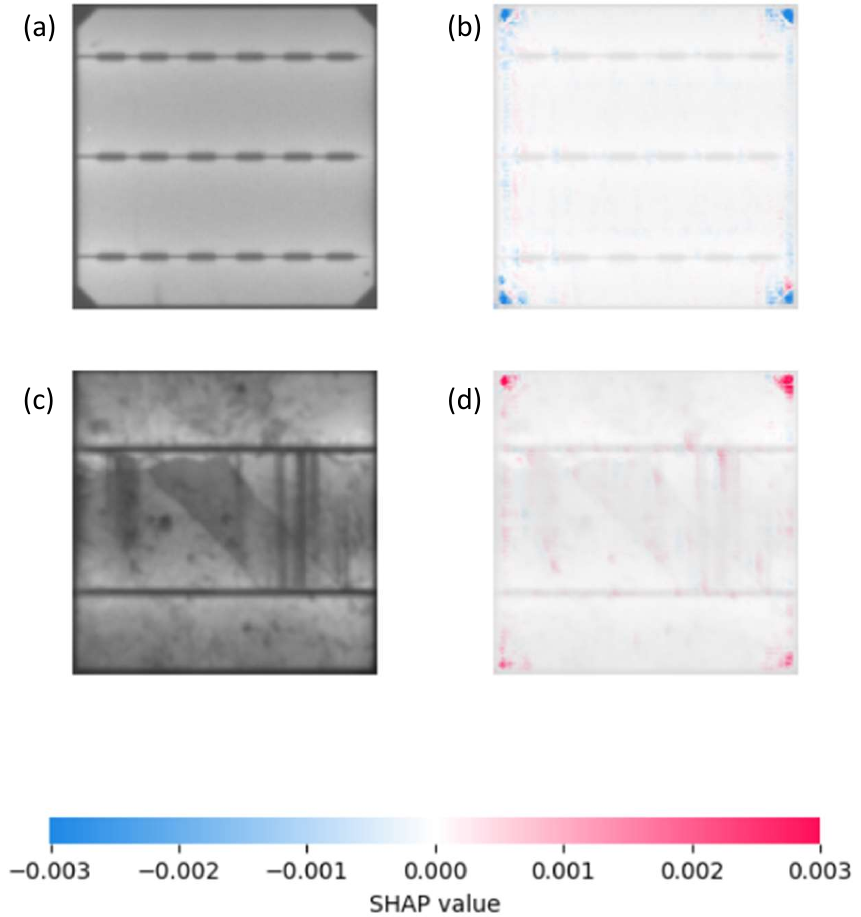}
    \caption{
    Images of (a) monocrystalline and (b) polycrystalline silicon solar modules. The corresponding SHAP values for classification of the images into polycrystalline modules are plotted in (c) and (d). The SHAP mask reveals the importance of features during the classification of the object by the model. Red signifies a higher SHAP score i.e. a positive influence of the feature for the classification into that class. On the other hand, the blue mask signifies a negative SHAP score i.e. the negative influence of that feature on the probability of classification into the class. The red mask seen in the corners of the module in (d) shows that the filled corners had a positive influence on classification into polycrystalline module. On the other hand, the the blue mask as seen in (b) shows the negative influence of the feature for classifying it as polycrystalline module. The main body of the image remains white with an SHAP value close to zero indicating no influence. This mask generated by SHAP scores aligns with the patterns observed for classifying polycrystalline and monocrystalline silicon modules.
    \label{fig:SHAP_ELPV}}
\end{figure}

We apply SHAP to the CNN model for classifying solar modules in the ELPV dataset. 
\autoref{fig:SHAP_ELPV} shows images of monocrystalline and polycrystalline solar cells together with their SHAP values. The plot of SHAP values clearly identifies the corners of the modules as the regions responsible for the classification decision. Specifically, in the case of polycrystalline samples, the red shading in the corners indicates a positive influence on the prediction. Conversely, for monocrystalline silicon, the blue-highlighted cutouts in the corners signify a negative influence, suggesting a contrasting impact on the prediction. 
The remaining region has minimal influence on the model's decision. Compared to LIME, SHAP provides better interpretability for polycrystalline silicon samples.

\subsection{Gradient-weighted Class Activation Mapping (Grad-CAM)}
Grad-CAM is an extension of the class activation mapping (CAM) technique used to visualize regions of the input that influence the predictions made by CNNs. It incorporates target class-specific gradient information flowing into the final convolutional layer of a CNN to produce a coarse localization map of important regions in the image \cite{Selvaraju_2019}. It is class-discriminative and isolates relevant input image regions considered by the model when making a prediction. 
Unlike LIME, it does not require re-training and can be applied to any CNN-based architecture \cite{Selvaraju2016GradCAMWD}.

\begin{figure}[h]
    \centering
    \includegraphics[width=\textwidth]{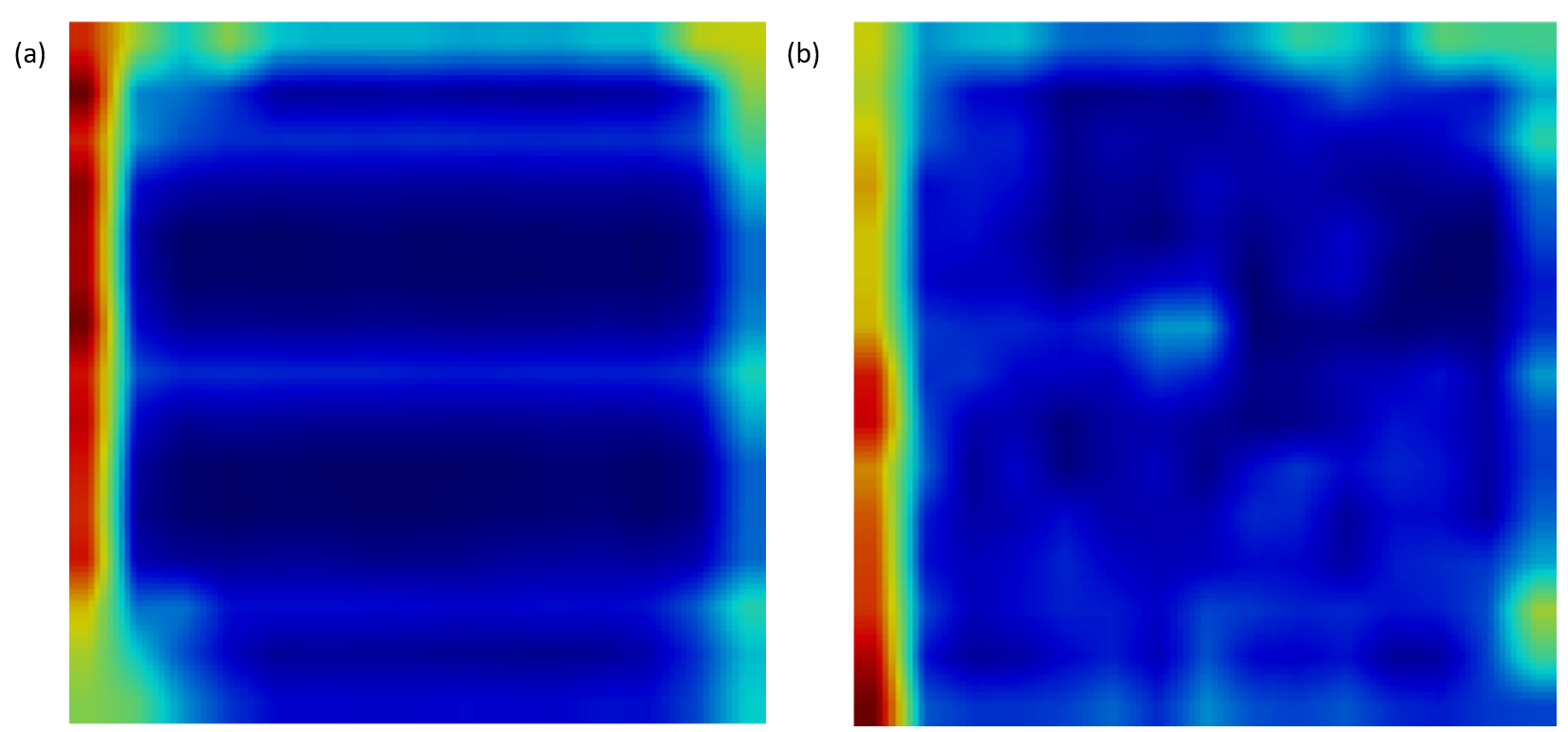}
    \caption{Application of Grad-CAM on (a) monocrystalline  \& (b) polycrystalline silicon module. The resulting mask is not as sharp as LIME or SHAP but still provides insights into the model's attention and focus. In addition, the interpretation of features such as the hotspot on the left edge is unclear.}
    \label{fig:gradcam_elpv}
\end{figure}

\autoref{fig:gradcam_elpv} shows the Grad-CAM mask applied to our CNN model for classifying modules into monocrystalline or polycrystalline classes. Grad-CAM falls short of offering a clear understanding of the model's focus. The model emphasizes the corners of the modules, but there also exists a prominent hot spot along the left boundary whose interpretation is unclear. This discrepancy in results, when compared to SHAP and LIME, stems from the distinct purposes of these techniques. Grad-CAM interprets inputs by showcasing areas of maximal activation. In contrast, both LIME and SHAP are employed to analyze the model's predictions. This difference in approach explains why Grad-CAM might highlight regions with high activation without directly correlating them to the model's decision-making process. The chosen interpretability technique should therefore align with the goals of the analysis.

\subsection{Interpretability analysis on PV defect classification}
\begin{figure}[h]
    \centering
    \includegraphics[width=0.7\textwidth]{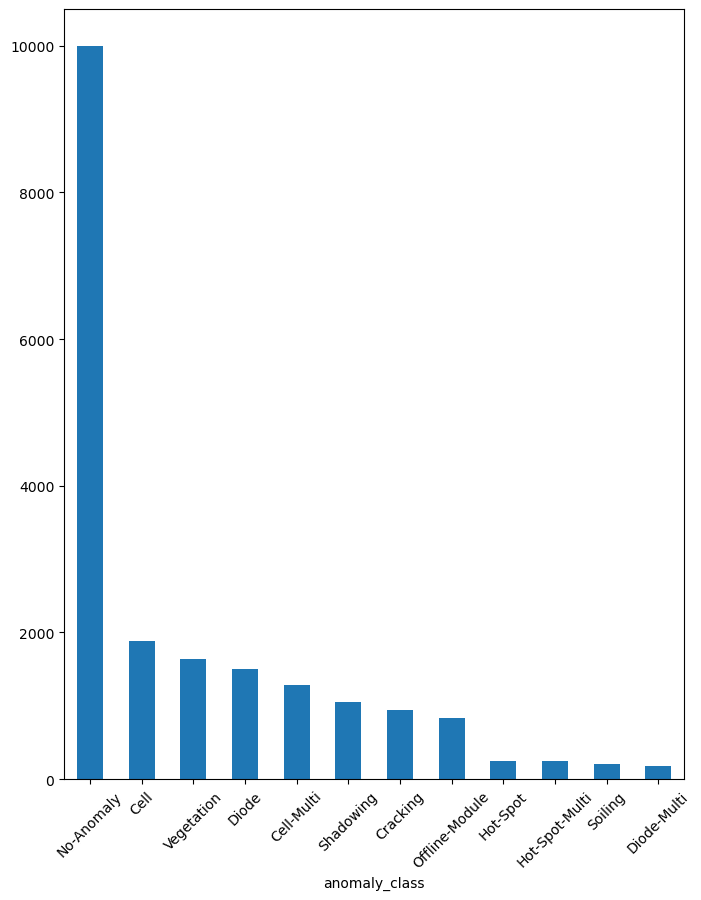}
    \caption{Frequency distribution of solar panel defect classes illustrating severe class imbalance.} 
    \label{fig:freq_dist}
\end{figure}

Next, we applied the interpretability analysis to the challenging problem of defect classification in PV modules. We employed the infrared solar module dataset for this task \cite{matthew_millendorf_edward_obropta_nikhil_vadhavkar_2023}.
Due to the imbalanced nature of the dataset as shown in \autoref{fig:freq_dist}, the synthetic minority oversampling technique was used to create a balanced training set. A custom CNN-based model consisting of 7 convolutional layers, with each layer followed by a max pooling and a batch normalization layer as shown in \autoref{cnn_final}, was used as the platform to demonstrate the interpretability. The model achieved a classification accuracy of 77\% on the test set. The low accuracy is due to the presence of several underrepresented classes. Removing the minority classes leads to better performance. To demonstrate the interpretability, we used the LIME method. SHAP was not used because the packaged implementation of SHAP in Python had some discrepancies in computing the Shapley values for the batch normalization layer.

\begin{figure}[htbp]
    \centering
    \includegraphics[width=\textwidth]{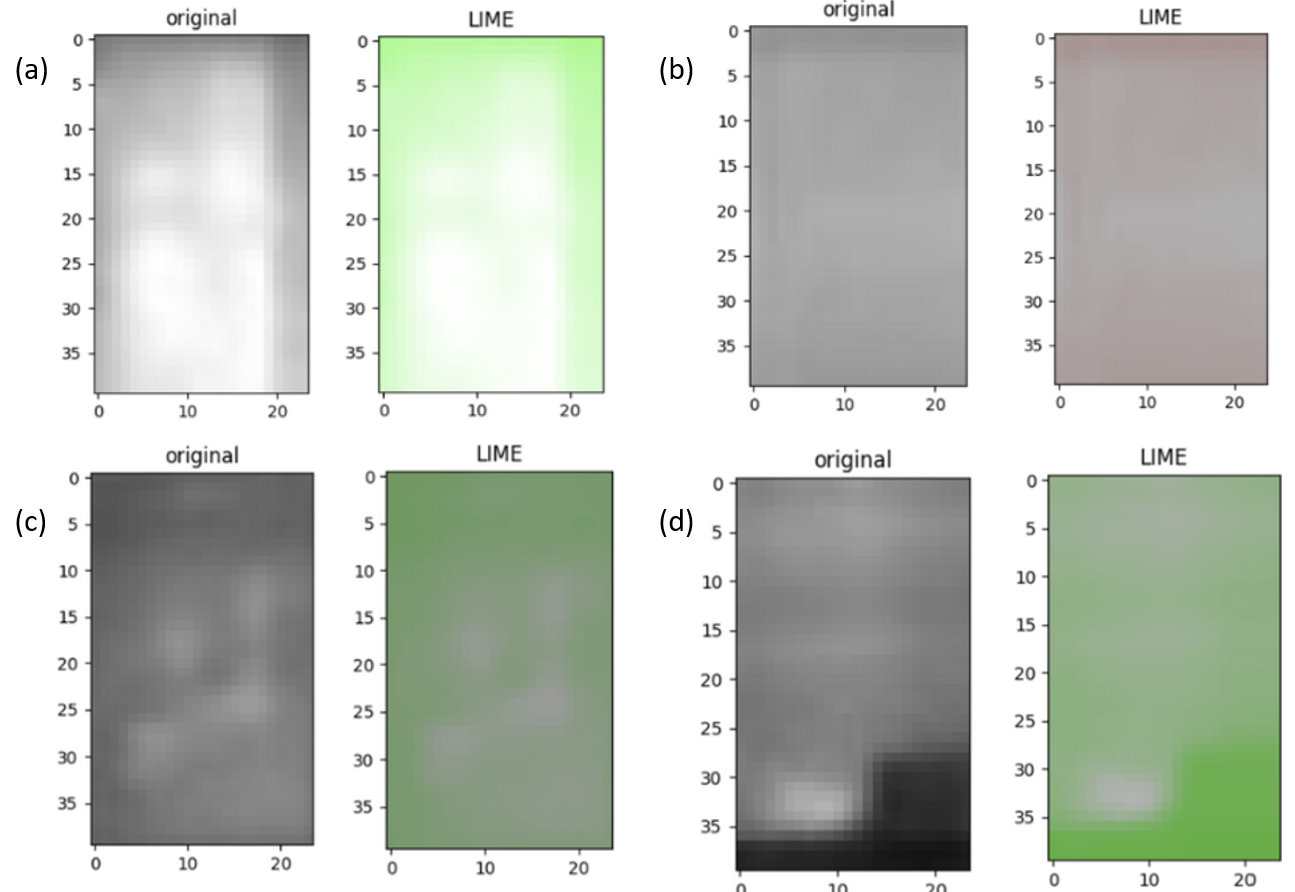}
    \caption{LIME masks for PV defect classification shown alongside the original IR image of the solar cell. (a) The original image shows a PV panel having a ``cell - multi defect" i.e. hot spots occurring with a square geometry in multiple cells.  along with the corresponding mask of accurate prediction. (b) The original image shows a PV panel having ``no  anomaly" along with the corresponding mask of accurate prediction. (c) The original image shows a PV panel having ``shadowing" along with the corresponding mask of wrong prediction, ``cracking". (d) The original image shows a PV panel having ``vegetation" along with the corresponding mask of accurate prediction. The LIME image mask reveals that the model predominantly focuses on the darker areas of the solar cells.}
    \label{fig: LIME pred masks}
\end{figure}

\begin{table}[htbp]
  \centering
  \footnotesize
  \caption{Neural network architecture used for PV defect classification to demonstrate interpretability analysis.}
  \label{cnn_final}
  \begin{tabular}{l c c c}
    \toprule
    \textbf{Layer (type)} & \textbf{Output Shape} & \textbf{Param \#} \\
    \midrule
    conv2d\_12 (Conv2D) & (None, 40, 24, 32) & 896 \\
    batch\_normalization\_6 & (None, 40, 24, 32) & 128 \\
    max\_pooling2d\_12 & (None, 20, 12, 32) & 0 \\
    conv2d\_13 (Conv2D) & (None, 20, 12, 32) & 9248 \\
    batch\_normalization\_7 & (None, 20, 12, 32) & 128 \\
    max\_pooling2d\_13 & (None, 10, 6, 32) & 0 \\
    conv2d\_14 (Conv2D) & (None, 10, 6, 64) & 18496 \\
    batch\_normalization\_8 & (None, 10, 6, 64) & 256 \\
    max\_pooling2d\_14 & (None, 5, 3, 64) & 0 \\
    conv2d\_15 (Conv2D) & (None, 5, 3, 128) & 73856 \\
    batch\_normalization\_9 & (None, 5, 3, 128) & 512 \\
    max\_pooling2d\_15 & (None, 3, 2, 128) & 0 \\
    conv2d\_16 (Conv2D) & (None, 3, 2, 256) & 295168 \\
    batch\_normalization\_10 & (None, 3, 2, 256) & 1024 \\
    max\_pooling2d\_16 & (None, 2, 1, 256) & 0 \\
    conv2d\_17 (Conv2D) & (None, 2, 1, 512) & 1180160 \\
    batch\_normalization\_11 &
 (None, 2, 1, 512) & 2048 \\
    max\_pooling2d\_17 & (None, 1, 1, 512) & 0 \\
    flatten\_2 (Flatten) & (None, 512) & 0 \\
    dense\_6 (Dense) & (None, 256) & 131328 \\
    dropout\_4 (Dropout) & (None, 256) & 0 \\
    dense\_7 (Dense) & (None, 128) & 32896 \\
    dropout\_5 (Dropout) & (None, 128) & 0 \\
    dense\_8 (Dense) & (None, 10) & 1290 \\
    \midrule
    \textbf{Total params:} & & 1747434 (6.67 MB) \\
    \textbf{Trainable params:} & & 1745386 (6.66 MB) \\
    \textbf{Non-trainable params:} & & 2048 (8.00 KB) \\
    \bottomrule
  \end{tabular}
\end{table}
\normalsize
The defects classified in this experiment are cell defects, vegetation growth, diode defects, shadowing, cracking, offline module defects, hot spot defects, and soiling. The defects appear in the images as either dark spots or light spots. By analyzing the various outputs generated by LIME, as shown in \autoref{fig: LIME pred masks}, alongside the model's performance, we can conclude that the model primarily focuses on the darker regions of the image for decision-making. The LIME masks, represented in green, intensify over the darker areas of the panel, while they appear lighter over the brighter regions, indicating that the model attributes more significance to these darker areas when making predictions. Therefore, this model will have difficulty in identifying defects that have intricate features within the brighter regions.
The main cause of this bias towards darker regions is that the defects that are associated with brighter regions such as hot spots and diode multi-short circuits have fewer data compared to the other defects as seen from \autoref{fig:freq_dist}. Hence, the interpretability analysis using LIME enables the identification of the flaws in the model, which can be mitigated by generating additional data for some classes.

\section{Outlook}
Reliable and robust techniques for detecting defects in solar cells are vital for maintaining high efficiency of PV plants over their life cycle to maximize their positive environmental impact. Existing defect detection methods face numerous challenges that can be primarily categorized into two areas: data-related issues and machine learning model limitations. Due to the various classes of defects with each having their own intricate features, a large dataset is required for training. Further, the dataset needs to be balanced with respect to all defect classes to distinguish them correctly. The collection of high quality data for training is capital and time-intensive process requiring investment in UAVs and cameras. Defect detection models may suffer from low accuracy and generalizability due to factors such as complex defect patterns and varying environmental conditions. Further, deep learning models require large computing power for training and inference.
We highlight important areas for further progress in achieving PV defect detection required for commercial-scale impact below.

During data collection, it is essential to ensure a degree of redundancy in images of the generated dataset. This means having multiple images of the same solar cell at different angles, conditions, and times. Multiple images of the same cell at different angles can be obtained by extracting them from the same video captured by UAV.
Further, a combination of different types of images such as IR, EL, and RGB are required since some defects are only visible in one type of image. Having redundancy and then performing a majority vote on the defect classification has been shown to improve the accuracy significantly. To mitigate data imbalance issue, the UAV trajectory can be designed to focus more on defects that are less prevalent in the dataset. The generation of synthetic images is another strategy for improving the classification accuracy of some classes.  In addition, instead of relying solely on experimental data that might not be representative enough, the geometric characteristic learning region-based CNN (GCL R-CNN) approach that utilizes artificially created samples to guide the training of real input samples has proven to be effective \cite{yongjie2022gcl_rcnn}. This approach enhances the model's awareness of the geometric structure of defective surfaces, thereby improving the accuracy of defect detection.

Numerous opportunities exist to enhance defect detection accuracy by choosing a machine learning model that leverages domain-specific knowledge and expertise. Existing models use data augmentation to encode rotation invariance for defect detection, eventually increasing the dataset size by 5-20x. This technique is very inefficient as obtaining an angular resolution of $\alpha$ in 2D requires $\mathcal{O}(\alpha^{-1})$ more filters in the model~\cite{thomasTensorFieldNetworks2018}. The large dataset increases the time required for training. Rotational-equivariant filters are a natural extension of translational-equivariant filters used in traditional CNNs used in computer vision. These filters provide extra information to the model about the properties under rotation while ensuring the final classification is rotation invariant. Harmonic neural networks leverage circular harmonic filters to ensure patchwise translation and 360$^{\circ}$-rotational equivariance, thereby enhancing the model's ability to generalize its performance and bolster its robustness \cite{harmonic-net-equivariance}. 
These networks achieve better accuracy than CNNs with augmented data over different training set sizes. 
These techniques fall under the umbrella of geometric deep learning (GDL) that is gaining traction in applications such as network analysis, recommender systems, particle physics, and chemistry~\cite{bronstein2021geometric}. GDL expands traditional deep learning techniques to accommodate non-Euclidean data like graphs and meshes. For detecting defects in solar cells, GDL may prove useful for capturing and analyzing spatial relationships and geometric structures in 3D images. The detection of invisible subsurface defects in PV modules is a challenging task when relying solely on 2D images, as they lack the necessary depth information. However, by incorporating additional geometrical information such as surface profile and 3D image of the PV modules, it may be possible to identify these hidden defects as well as anomalies such as bent modules \cite{od_subsurface}.
Advanced data augmentation techniques that rely on geometric transformation applied to defect only and not the entire image can boost the accuracy of defect classification of semiconductor wafers~\cite{dataAug-geotrans}.

Physics-informed neural networks (PINNs) present opportunities to incorporate physical laws that describe the formation and evolution of defects in solar cells, thereby enabling the detection of defects not easily identified in images.
These networks are particularly useful in IR image-based defect detection. By integrating the governing partial differential equations of physical laws that describe heat diffusion and material properties across defective surfaces, PINNs can facilitate an accurate reconstruction of background information from IR images. \cite{pinn_thermal}. 
The training of robust machine learning models for defect detection in PV modules is hindered by the lack of large datasets featuring defective surfaces. To address this challenge, a physics-guided approach can be employed to generate photo-realistic synthetic images by solving the relevant physical laws, which can be combined with real images to enhance defect detection accuracy and promote model generalization \cite{synthetic_images_pinn}. The model can  be subsequently fine-tuned through transfer learning to classify defects in realistic scenarios ~\cite{Yousefian2023ImprovedPF}, thereby improving the model's effectiveness in detecting defects in PV modules. Finally, the model predictions need to be carefully examined through interpretability analyses described above to identify the causes of incorrect classification and areas of improvement.

\section{Data availability}
The code and data for interpretability analyses can be found on \href{https://github.com/sway-am/Interpretability-of-CNN-during-PV-classification}{GitHub}.

\section*{Acknowledgements}
ZA thanks the Texas Tech University Mechanical Engineering department startup grant for support of this research. VS is the founder of OnePV and Vayv Energy Systems. The research was conducted independent of the author's business interests.

\section*{References}
\bibliography{refs,ML}

\providecommand{\noopsort}[1]{}\providecommand{\singleletter}[1]{#1}%
\providecommand{\newblock}{}
\begin{thebibliography}{10}
\expandafter\ifx\csname url\endcsname\relax
  \def\url#1{{\tt #1}}\fi
\expandafter\ifx\csname urlprefix\endcsname\relax\def\urlprefix{URL }\fi
\providecommand{\eprint}[2][]{\url{#2}}

\bibitem{PVmarket}
{International Energy Agency} 2024 Snapshot of global pv markets
  \url{https://iea-pvps.org/wp-content/uploads/2024/04/Snapshot-of-Global-PV-Markets-1.pdf}
  accessed: 2024-06-23

\bibitem{marketsize}
{} 2024 Global solar power market size, share, trends, growth: By technology:
  Solar photovoltaic, concentrated solar power; by application: Residential,
  non-residential, utility; regional analysis; market dynamics: Swot analysis,
  porter’s five forces analysis; competitive landscape; key trends and
  developments in the market; 2024-2032 source:
  https://www.expertmarketresearch.com/reports/solar-power-market
  \url{https://www.expertmarketresearch.com/reports/solar-power-market}
  accessed: 2024-06-23

\bibitem{saur}
{Saur Energy} 2024 Solar o\&m market set to grow to \$15 billion by
  2030-woodmac
  \url{https://www.saurenergy.com/solar-energy-news/solar-om-market-set-to-grow-to-15-billion-by-2030-woodmac}
  accessed: 2024-06-23

\bibitem{Technology}
{Thomas Bruckner and Lew Fulton and Edgar Hertwich and Alan McKinnon and Daniel
  Perczyk and Joyashree Roy and Roberto Schaeffer and Steffen Schlömer and
  Ralph Sims and Pete Smith and Ryan Wise} 2024 Technology-specific cost and
  performance parameters
  \url{https://www.ipcc.ch/site/assets/uploads/2018/02/ipcc_wg3_ar5_annex-iii.pdf}
  accessed: 2024-06-23

\bibitem{wind}
{Wind Energy Technologies Office} 2024 How wind can help us breathe easier
  \url{https://www.energy.gov/eere/wind/articles/how-wind-can-help-us-breathe-easier#:~:text=CO2%20Emissions%20from%20Different%20Energy%20Sources&text=Wind%20energy%20produces%20around%2011,2%2FkWh%20for%20natural%20gas.}
  accessed: 2024-06-23

\bibitem{bartlerAutomatedDetectionSolar2018}
Bartler A, Mauch L, Yang B, Reuter M and Stoicescu L 2018 Automated
  {{Detection}} of {{Solar Cell Defects}} with {{Deep Learning}} {\em 2018 26th
  {{European Signal Processing Conference}} ({{EUSIPCO}})\/} (Rome: IEEE) pp
  2035--2039 ISBN 978-90-827970-1-5

\bibitem{TANG2020453}
Tang W, Yang Q, Xiong K and Yan W 2020 {\em Solar Energy\/} {\bf 201} 453--460
  ISSN 0038-092X
  \urlprefix\url{https://www.sciencedirect.com/science/article/pii/S0038092X20302875}

\bibitem{chenAutomatedDefectIdentification2022}
Chen X, Karin T and Jain A 2022 {\em Solar Energy\/} {\bf 242} 20--29 ISSN
  0038092X

\bibitem{hijjawiReviewAutomatedSolar2023}
Hijjawi U, Lakshminarayana S, Xu T, Piero Malfense~Fierro G and Rahman M 2023
  {\em Solar Energy\/} {\bf 266} 112186 ISSN 0038092X

\bibitem{Salameh2020LifeCA}
Salameh T, Tawalbeh M, Alami A~H, Al-Othman A, Issa S and Alkasrawi M 2020 {\em
  2020 Advances in Science and Engineering Technology International Conferences
  (ASET)\/}  1--6
  \urlprefix\url{https://api.semanticscholar.org/CorpusID:219858502}

\bibitem{rs15061686}
Boubaker S, Kamel S, Ghazouani N and Mellit A 2023 {\em Remote Sensing\/} {\bf
  15} ISSN 2072-4292 \urlprefix\url{https://www.mdpi.com/2072-4292/15/6/1686}

\bibitem{SINGH201236}
Singh P and Ravindra N 2012 {\em Solar Energy Materials and Solar Cells\/} {\bf
  101} 36--45 ISSN 0927-0248
  \urlprefix\url{https://www.sciencedirect.com/science/article/pii/S0927024812000931}

\bibitem{Kauppinen2015}
Kauppinen T, Panouillot P~E, Siikanen S, Athanasakou E, Baltas P and
  Nikopoulous B 2015 About infrared scanning of photovoltaic solar plant {\em
  Thermosense: Thermal Infrared Applications XXXVII\/} vol 9485 ed Hsieh S~J~T
  and Zalameda J~N (SPIE) p 948517 ISSN 0277-786X
  \urlprefix\url{http://dx.doi.org/10.1117/12.2180165}

\bibitem{Dhimish2021}
Dhimish M and Lazaridis P~I 2021 {\em Scientific Reports\/} {\bf 11} 23961 ISSN
  2045-2322 \urlprefix\url{https://doi.org/10.1038/s41598-021-03498-z}

\bibitem{article}
Nieto-Vallejo A~E, Ruiz F and Patino D 2019 {\em DYNA\/} {\bf 86} 54--63

\bibitem{solar3020019}
Badran G and Dhimish M 2023 {\em Solar\/} {\bf 3} 322--346 ISSN 2673-9941
  \urlprefix\url{https://www.mdpi.com/2673-9941/3/2/19}

\bibitem{Meng}
Meng Z, Xu S, Wang L, Gong Y, Zhang X and Zhao Y 2022 {\em Energy Science \&
  Engineering\/} {\bf 10} 800--813 (\textit{Preprint}
  \eprint{https://onlinelibrary.wiley.com/doi/pdf/10.1002/ese3.1056})
  \urlprefix\url{https://onlinelibrary.wiley.com/doi/abs/10.1002/ese3.1056}

\bibitem{hoyer2010electroluminescence}
Hoyer U, Wagner M, Swonke T, Bachmann J, Auer R, Osvet A and Brabec C~J 2010
  {\em Applied Physics Letters\/} {\bf 97} 233303 ISSN 1077-3118
  \urlprefix\url{http://dx.doi.org/10.1063/1.3521259}

\bibitem{[3]}
Pratt L, Govender D and Klein R 2021 {\em Renewable Energy\/} {\bf 178}
  1211--1222 ISSN 0960-1481

\bibitem{[2]}
A~Vageswar K~B and Krishnamurthy C 2010 {\em Nondestructive Testing and
  Evaluation\/} {\bf 25} 333--340

\bibitem{8733881}
Dhimish M and Mather P 2019 {\em IEEE Transactions on Semiconductor
  Manufacturing\/} {\bf 32} 277--285

\bibitem{8016569}
Zafirovska I, Juhl M~K, Weber J~W, Wong J and Trupke T 2017 {\em IEEE Journal
  of Photovoltaics\/} {\bf 7} 1496--1502

\bibitem{rahaman2022infrared}
Rahaman S~A 2022 {\em Infrared (IR) Imaging-based Inspection for Performance
  Measurements of Solar PV Systems\/} Ph.D. thesis Murdoch University

\bibitem{8478340}
Li X, Yang Q, Lou Z and Yan W 2019 {\em IEEE Transactions on Energy
  Conversion\/} {\bf 34} 520--529

\bibitem{Pern1991}
Pern F, Czanderna A, Emery K and Dhere R 1991 Weathering degradation of eva
  encapsulant and the effect of its yellowing on solar cell efficiency {\em
  Proc. 22nd IEEE Photovoltaic Specialists Conference\/} vol~1 pp 557 -- 561
  ISBN 0-87942-636-5

\bibitem{Segbefia2023-ym}
Segbefia O~K 2023 {\em Heliyon\/} {\bf 9} e19566 ISSN 2405-8440
  \urlprefix\url{http://dx.doi.org/10.1016/j.heliyon.2023.e19566}

\bibitem{Wohlgemuth2015}
Wohlgemuth J, Silverman T, Miller D~C, McNutt P, Kempe M and Deceglie M 2015
  Evaluation of pv module field performance {\em 2015 IEEE 42nd Photovoltaic
  Specialist Conference (PVSC)\/} vol~16 (IEEE) p 1–7
  \urlprefix\url{http://dx.doi.org/10.1109/PVSC.2015.7356132}

\bibitem{Liu2015}
Liu H~C, Huang C~T, Lee W~K, Yan S~S and Lin F~M 2015 {\em Energy and Power
  Engineering\/} {\bf 07} 348–353 ISSN 1947-3818
  \urlprefix\url{http://dx.doi.org/10.4236/epe.2015.78032}

\bibitem{Abdelaziz2022}
Abdelaziz G, Hichem H, Chiheb B~R and Rached G 2021 Shading effect on the
  performance of a photovoltaic panel {\em 2021 IEEE 2nd International
  Conference on Signal, Control and Communication (SCC)\/} vol~5 (IEEE) p
  208–213 \urlprefix\url{http://dx.doi.org/10.1109/SCC53769.2021.9768356}

\bibitem{mahmoud_meribout__2023}
Meribout M, Tiwari V~K, Herrera J~P~P and Baobaid A~N~M~A 2023 {\em
  Measurement\/} {\bf 209} 112466--112466

\bibitem{https://doi.org/10.1002/pip.3448}
Bommes L, Pickel T, Buerhop-Lutz C, Hauch J, Brabec C and Peters I~M 2021 {\em
  Progress in Photovoltaics: Research and Applications\/} {\bf 29} 1236--1251
  (\textit{Preprint}
  \eprint{https://onlinelibrary.wiley.com/doi/pdf/10.1002/pip.3448})
  \urlprefix\url{https://onlinelibrary.wiley.com/doi/abs/10.1002/pip.3448}

\bibitem{NAVEENVENKATESH2022110786}
{Naveen Venkatesh} S and Sugumaran V 2022 {\em Measurement\/} {\bf 191} 110786
  ISSN 0263-2241
  \urlprefix\url{https://www.sciencedirect.com/science/article/pii/S0263224122000860}

\bibitem{gedraite2011investigation}
Gedraite E~S and Hadad M 2011 Investigation on the effect of a gaussian blur in
  image filtering and segmentation {\em Proceedings ELMAR-2011\/} pp 393--396

\bibitem{10.1007/978-3-319-10602-1_48}
Lin T~Y, Maire M, Belongie S, Hays J, Perona P, Ramanan D, Doll{\'a}r P and
  Zitnick C~L 2014 Microsoft coco: Common objects in context {\em Computer
  Vision -- ECCV 2014\/} ed Fleet D, Pajdla T, Schiele B and Tuytelaars T
  (Cham: Springer International Publishing) pp 740--755 ISBN 978-3-319-10602-1

\bibitem{Patel}
Patel A~V, McLauchlan L and Mehrubeoglu M 2020 Defect detection in pv arrays
  using image processing {\em 2020 International Conference on Computational
  Science and Computational Intelligence (CSCI)\/} (IEEE) p 1653–1657
  \urlprefix\url{http://dx.doi.org/10.1109/CSCI51800.2020.00304}

\bibitem{bayesianNN}
Kang G, Gao S, Yu L, Zhang D, Wei X and Zhan D 2019 {\em IEEE Access\/} {\bf 7}
  173366–173376 ISSN 2169-3536
  \urlprefix\url{http://dx.doi.org/10.1109/ACCESS.2019.2955753}

\bibitem{harmonic-net-equivariance}
Worrall D~E, Garbin S~J, Turmukhambetov D and Brostow G~J 2017 Harmonic
  networks: Deep translation and rotation equivariance {\em 2017 IEEE
  Conference on Computer Vision and Pattern Recognition (CVPR)\/} (IEEE)
  \urlprefix\url{http://dx.doi.org/10.1109/CVPR.2017.758}

\bibitem{devries2019evaluation}
DeVries T, Romero A, Pineda L, Taylor G~W and Drozdzal M 2019 {\em arXiv
  preprint arXiv:1907.08175\/}

\bibitem{barua2019qualityevaluationgansusing}
Barua S, Ma X, Erfani S~M, Houle M~E and Bailey J 2019 {\em arXiv preprint
  arXiv:1905.00643\/}

\bibitem{gulrajani2017improved}
Gulrajani I, Ahmed F, Arjovsky M, Dumoulin V and Courville A~C 2017 {\em
  Advances in neural information processing systems\/} {\bf 30}

\bibitem{Deitsch_2019}
Deitsch S, Christlein V, Berger S, Buerhop-Lutz C, Maier A, Gallwitz F and
  Riess C 2019 {\em Solar Energy\/} {\bf 185} 455–468 ISSN 0038-092X
  \urlprefix\url{http://dx.doi.org/10.1016/j.solener.2019.02.067}

\bibitem{inproceedings}
Zyout I and Oatawneh A 2020 Detection of pv solar panel surface defects using
  transfer learning of the deep convolutional neural networks {\em 2020
  Advances in Science and Engineering Technology International Conferences
  (ASET)\/} (IEEE) pp 1--4
  \urlprefix\url{http://dx.doi.org/10.1109/ASET48392.2020.9118382}

\bibitem{9398560}
Su B, Chen H and Zhou Z 2022 {\em IEEE Transactions on Industrial
  Electronics\/} {\bf 69} 3161--3171

\bibitem{Chen2020-fc}
Chen H, Hu Q, Zhai B, Chen H and Liu K 2020 {\em Neural Comput. Appl.\/} {\bf
  32} 11229--11244

\bibitem{Bilal_2018}
Bilal A, Jourabloo A, Ye M, Liu X and Ren L 2018 {\em IEEE Transactions on
  Visualization and Computer Graphics\/} {\bf 24} 152–162 ISSN 1077-2626
  \urlprefix\url{http://dx.doi.org/10.1109/TVCG.2017.2744683}

\bibitem{PRATT2023200048}
Pratt L, Mattheus J and Klein R 2023 {\em Systems and Soft Computing\/} {\bf 5}
  200048 ISSN 2772-9419
  \urlprefix\url{https://www.sciencedirect.com/science/article/pii/S2772941923000017}

\bibitem{guptasegmentation2020}
Gupta D 2020 {\em GitHub repository (accessed 2024-09-02)\/}
  \urlprefix\url{https://github.com/divamgupta/image- segmentation-keras}

\bibitem{9301240}
Ge C, Liu Z, Fang L, Ling H, Zhang A and Yin C 2021 {\em IEEE Transactions on
  Parallel and Distributed Systems\/} {\bf 32} 1653--1664

\bibitem{liang2021explaining}
Liang Y, Li S, Yan C, Li M and Jiang C 2021 {\em Neurocomputing\/} {\bf 419}
  168--182

\bibitem{Buerhop2018}
Buerhop-Lutz C, Deitsch S, Maier A, Gallwitz F, Berger S, Doll B, Hauch J,
  Camus C and Brabec C~J 2018 A benchmark for visual identification of
  defective solar cells in electroluminescence imagery {\em 35th European PV
  Solar Energy Conference and Exhibition\/} vol 12871289 pp 1287--1289

\bibitem{Deitsch2021}
Deitsch S, Buerhop-Lutz C, Sovetkin E, Steland A, Maier A, Gallwitz F and Riess
  C 2021 {\em Machine Vision and Applications\/} {\bf 32} 84
  \urlprefix\url{http://dx.doi.org/10.1007/s00138-021-01191-9}

\bibitem{ribeiro2016why}
Ribeiro M~T, Singh S and Guestrin C 2016 "why should i trust you?": Explaining
  the predictions of any classifier {\em Proceedings of the 22nd ACM SIGKDD
  International Conference on Knowledge Discovery and Data Mining\/} KDD '16
  (New York, NY, USA: Association for Computing Machinery) p 1135–1144 ISBN
  9781450342322
  \urlprefix\url{https://doi-org.lib-e2.lib.ttu.edu/10.1145/2939672.2939778}

\bibitem{faigle1992shapley}
Faigle U and Kern W 1992 {\em International Journal of Game Theory\/} {\bf 21}
  249--266

\bibitem{mitra2023novel}
Mitra S and Gilpin L 2023 {\em arXiv preprint arXiv:2311.10811\/}

\bibitem{ledel2022studying}
Schulte L, Ledel B and Herbold S 2024 {\em Empirical Software Engineering\/}
  {\bf 29} ISSN 1573-7616
  \urlprefix\url{http://dx.doi.org/10.1007/s10664-024-10469-1}

\bibitem{Selvaraju_2019}
Selvaraju R~R, Cogswell M, Das A, Vedantam R, Parikh D and Batra D 2017
  Grad-cam: Visual explanations from deep networks via gradient-based
  localization {\em 2017 IEEE International Conference on Computer Vision
  (ICCV)\/} pp 618--626

\bibitem{Selvaraju2016GradCAMWD}
Selvaraju R~R, Cogswell M, Das A, Vedantam R, Parikh D and Batra D 2019 {\em
  International Journal of Computer Vision\/} {\bf 128} 336–359 ISSN
  1573-1405 \urlprefix\url{http://dx.doi.org/10.1007/s11263-019-01228-7}

\bibitem{matthew_millendorf_edward_obropta_nikhil_vadhavkar_2023}
Millendorf M, Obropta E and Vadhavkar N 2023 Infrared solar modules
  \urlprefix\url{https://www.kaggle.com/dsv/6866566}

\bibitem{yongjie2022gcl_rcnn}
Zhai Y, Yang K, Zhao Z, Wang Q and Bai K 2022 {\em Engineering Applications of
  Artificial Intelligence\/} {\bf 116} 105429 ISSN 0952-1976
  \urlprefix\url{http://dx.doi.org/10.1016/j.engappai.2022.105429}

\bibitem{thomasTensorFieldNetworks2018}
Thomas N, Smidt T, Kearnes S, Yang L, Li L, Kohlhoff K and Riley P 2018 {\em
  arXiv:1802.08219 [cs]\/} (\textit{Preprint} \eprint{1802.08219})

\bibitem{bronstein2021geometric}
Bronstein M~M, Bruna J, Cohen T and Veli{\v{c}}kovi{\'c} P 2021 {\em arXiv
  preprint arXiv:2104.13478\/}

\bibitem{od_subsurface}
Timilsina S, Jang S~M, Jo C~W, Kwon Y~N, Sohn K~S, Lee K~H and Kim J~S 2023
  {\em Advanced Intelligent Systems\/} {\bf 5} 2300314

\bibitem{dataAug-geotrans}
{López de la Rosa} F, Gómez-Sirvent J~L, Morales R, Sánchez-Reolid R and
  Fernández-Caballero A 2023 {\em Computers \& Industrial Engineering\/} {\bf
  183} 109549 ISSN 0360-8352
  \urlprefix\url{https://www.sciencedirect.com/science/article/pii/S0360835223005739}

\bibitem{pinn_thermal}
Lim W~H, Sfarra S and Yao Y 2021 {\em Engineering Proceedings\/} {\bf 8} 14
  ISSN 2673-4591 \urlprefix\url{https://www.mdpi.com/2673-4591/8/1/14}

\bibitem{synthetic_images_pinn}
Manyar O~M, Cheng J, Levine R, Krishnan V, Barbi{\v{c}} J and Gupta S~K 2023
  {\em Journal of Computing and Information Science in Engineering\/} {\bf 23}
  030903

\bibitem{Yousefian2023ImprovedPF}
Yousefian P, Sepehrinezhad A, van Duin A~C~T and Randall C~A 2023 {\em APL
  Machine Learning\/} {\bf 1} 036107

\end{thebibliography}
\end{document}